\documentclass{article}
\usepackage{jfrExamplee}
\usepackage{graphicx}
\usepackage{apalike}
\usepackage{setspace}
\usepackage{authblk}
\usepackage{subfigure}
\usepackage{cite}
\usepackage{amsmath,amssymb,amsfonts}
\usepackage{algorithmic}
\usepackage{bbding}
\usepackage{textcomp}
\usepackage{bm}
\usepackage{float}
\usepackage{diagbox}
\usepackage{multirow}
\usepackage{harpoon}
\usepackage{stfloats}
\usepackage{float}
\usepackage{booktabs}
\usepackage{caption,setspace}
\usepackage{multirow}
\usepackage{changes}
\usepackage{url}

\usepackage[edges]{forest}
\usepackage[capitalize]{cleveref}

\newsavebox\CBox

\makeatletter
\def \thanks#1{{\protected@xdef\@thanks{\@thanks\protect\footnotetext{#1}}}}
\makeatother

\title{Autonomous Driving in Unstructured Environments: How Far Have We Come?}

\author{ \\Chen Min\textsuperscript{*}, Shubin Si\textsuperscript{*}, Xu Wang\textsuperscript{*}, Hanzhang Xue\textsuperscript{*}, Weizhong Jiang\textsuperscript{*},  Zitong Chen, Mengmeng Li, \\Jilin Mei, Erke Shang, Zhipeng Xiao, Bin Dai, Qi Zhu, Hao Fu, Dawei Zhao, Liang Xiao, Yiming Nie\textsuperscript{\dag}, \\ Yu Hu\textsuperscript{\dag}
	\thanks{Chen Min, Jilin Mei, and Yu Hu are with the Research Center for Intelligent Computing Systems, SKLP, Institute of Computing Technology, Chinese Academy of Sciences, Beijing, China. email:{\tt\small \{mincheng, meijilin, huyu\}@ict.ac.cn}}
	\thanks{Shubin Si is with Harbin Engineering University, Harbin, China. email:{\tt\small sishubin@hrbeu.edu.cn}}
	\thanks{Xu Wang is with Jianghuai Advance Technology Center, Anhui Provincial Key Laboratory of Humanoid Robot, Anhui Provincial Industry Innovation Center of Humanoid Robot, Hefei, Anhui, China. email:{\tt\small wangxucnah@163.com}}
	\thanks{Zitong Chen is with Beihang University, Beijing, China. email:{\tt\small zitong\_chen@buaa.edu.cn}} 
	\thanks{Hanzhang Xue is with Test Center, National University of Defense Technology, Xi'an, Shaanxi, China. email:{\tt\small xuehanzhang13@nudt.edu.cn}}
	\thanks{Hao Fu is with National University of Defense Technology, Changsha, China. email:{\tt\small fuhao@nudt.edu.cn}}
	\thanks{Weizhong Jiang, Mengmeng Li, Erke Shang, Zhipeng Xiao, Qi Zhu, Bin Dai, Dawei Zhao, Liang Xiao, and Yiming Nie are with the Unmanned Systems Technology Research Center, Defense Innovation Institute, Beijing, China. email: \tt\small erke1984@qq.com, \tt\small xiaozhipeng.cs@hotmail.com, \tt\small xiaoliang@nudt.edu.cn, \tt\small nieym@alumni.nudt.edu.cn, {\tt\small \{jwz\_0911, lmmbit, Zhuqi, ibindai, adamzdw\}@163.com}}
	\thanks{\textsuperscript{*}: Equal Contribution. \textsuperscript{\dag}: Corresponding Authors.
	}}

\begin{document}
	
\maketitle
	
\begin{abstract}
Research on autonomous driving in unstructured outdoor environments is less advanced than in structured urban settings due to challenges like environmental diversities and scene complexity. These environments—such as rural areas and rugged terrains—pose unique obstacles that are not common in structured urban areas. Despite these difficulties, autonomous driving in unstructured outdoor environments is crucial for applications in agriculture, mining, and military operations. Our survey reviews over 250 papers for autonomous driving in unstructured outdoor environments, covering offline mapping, pose estimation, environmental perception, path planning, end-to-end autonomous driving, datasets, and relevant challenges. We also discuss emerging trends and future research directions. This review aims to consolidate knowledge and encourage further research for autonomous driving in unstructured environments. To support ongoing work, we maintain an active repository with up-to-date literature and open-source projects at: \url{https://github.com/chaytonmin/Survey-Autonomous-Driving-in-Unstructured-Environments}.
\end{abstract}

In recent years, autonomous driving technology has wittnessed remarkable strides, particularly in structured environments such as urban areas and highways. Numerous autonomous driving systems have proven their capabilities in practical applications, thanks to their reliance on standardized markers like lane lines, traffic signals, and road signs. These markers enable autonomous vehicles to execute path planning and decision-making with substantial accuracy. However, the real world often presents environments that are not fully structured. 
Unstructured outdoor environments differ significantly from their structured urban counterparts. These unstructured environments like mountains, jungles, and deserts, lack clear road signs, traffic signals, and lane markings, and can include rural roads, off-road terrain, construction sites, parking lots, and other areas characterized by high complexity and unpredictability.


\begin{figure*} [t]
	\centering
	\includegraphics[width=0.95\textwidth]{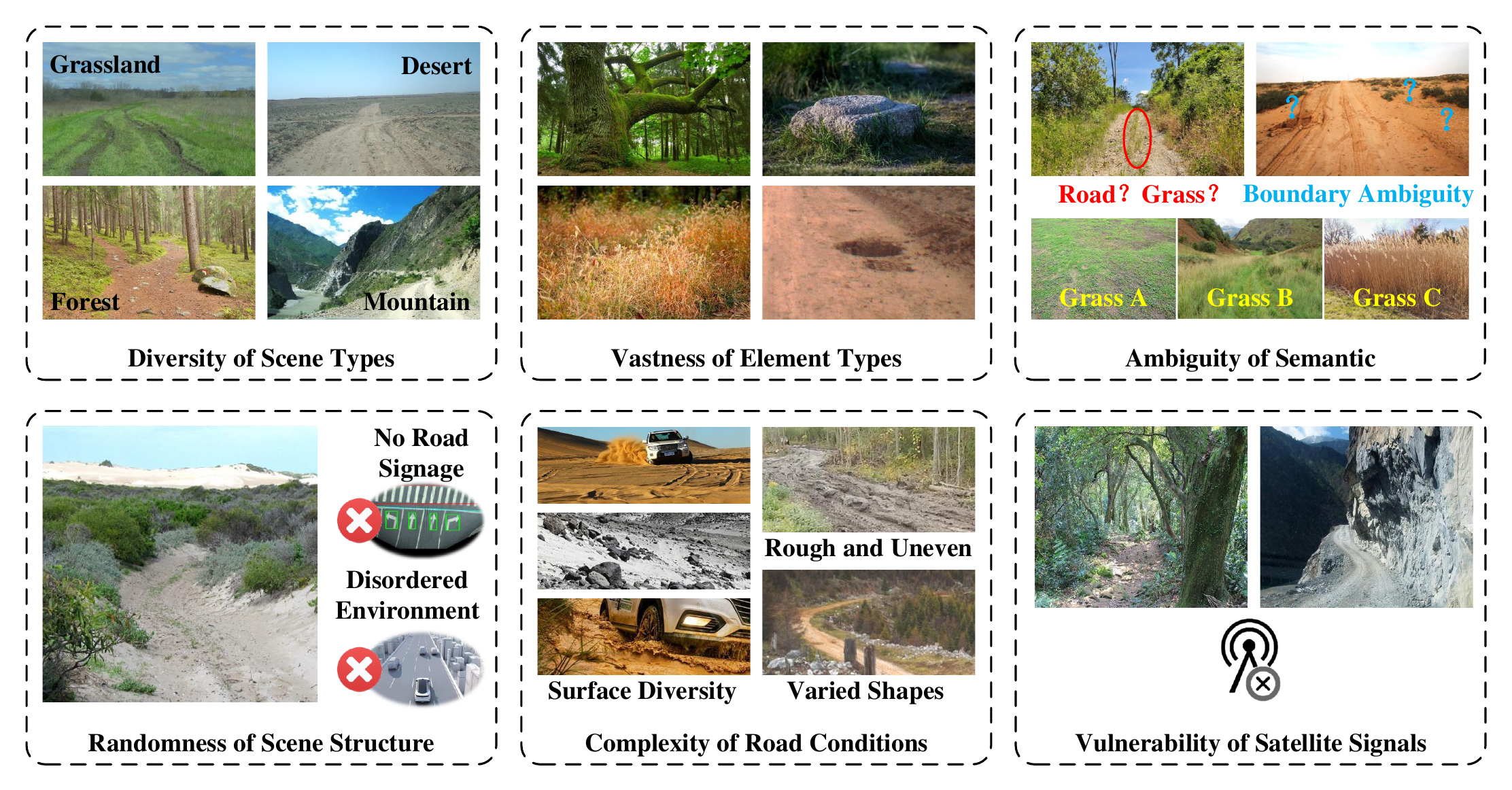}
	\caption{Inherent characteristics of unstructured environments. Compared to structured environments, autonomous driving in unstructured environment encounters various challenges, including the diversity of environmental types, a vast array of elements, ambiguity in semantic categories, disorderly scene structures, complex road conditions, and vulnerability of satellite signals.}
	\label{fig:features}
\end{figure*}

\tikzstyle{leaf}=[draw=black,
rounded corners,minimum height=1.2em,
text opacity=1, align=center,
fill opacity=.5,  text=black,align=left,font=\scriptsize,
inner xsep=3pt,
inner ysep=1pt,
]
\begin{figure*}[t]
	\centering
	\begin{forest}
		for tree={
			forked edges,
			grow=east,
			reversed=true,
			anchor=base west,
			parent anchor=east,
			child anchor=west,
			base=middle,
			font=\footnotesize,
			rectangle,
			draw=black,
			rounded corners,align=left,
			minimum width=2.5em,
			minimum height=1.2em,
			s sep=6pt,
			inner xsep=3pt,
			inner ysep=1pt,
		},
		where level=1{text width=4.5em}{},
		where level=2{text width=8em,font=\scriptsize}{},
		where level=3{font=\scriptsize}{},
		where level=4{font=\scriptsize}{},
		where level=5{font=\scriptsize}{},
		where level=6{font=\scriptsize}{},
		[\rotatebox{90}{\textbf{Autonomous Driving (AD) in Unstructured Environments  }}
		[Modular AD, text width=5.7 em
		[Offline Mapping,text width=9 em
		[LiDAR-based, text width=5.6 em
		[\cite{kaess2008isam}{,} \cite{pan2020gem}{,} ...,leaf,text width=16.8em]
		]
		[Fusion-based, text width=5.6 em
		[\cite{ren2021towards}{,} \cite{ren2021lidar}{,} ...,leaf,text width=16.8em]
		]
		]
		[Pose Estimation, text width=9 em
		[Matching-based, text width=5.6 em
		[\cite{peng2022roll}{,} \cite{li2020localization}{,} ...,leaf,text width=16.8em]
		]
		[Odometry-based, text width=5.6 em
		[\cite{marks2009gamma}{,} \cite{zhang2014loam}{,} ...,leaf,text width=16.8em]
		]
		]
		[Environmental Perception,text width=9 em
		[Traversability, text width=5.6 em
		[LiDAR-based,text width=8.5 em
		[\cite{chen2014gaussian},leaf,text width=6.6 em]
		]
		[Vision-based,text width=8.5 em
		[\cite{gao2021fine},leaf,text width=6.6 em]
		]
		[Fusion-based,text width=8.5 em
		[\cite{yan2024fsn},leaf,text width=6.6 em]
		]
		]
		[Segmentation, text width=5.6 em
		[LiDAR-based,text width=8.5 em
		[\cite{liu2021point},leaf,text width=6.6 em]
		]
		[Vision-based,text width=8.5 em
		[\cite{singh2021offroadtranseg},leaf,text width=6.6 em]
		]
		[Fusion-based,text width=8.5 em
		[\cite{feng2024multi},leaf,text width=6.6 em]
		]
		]
		]
		[Path Planning,text width=9em
		[Global,text width=5.6 em
		[Search-based, text width=8.5em
		[\cite{stentz1994optimal},leaf,text width=6.6 em]
		]
		[Sampling-based, text width=8.5em
		[\cite{jaillet2010sampling},leaf,text width=6.6 em]
		]
		[Biologically-inspired, text width=8.5em
		[\cite{wang2018off},leaf,text width=6.6 em]
		]
		]
		[Local,text width=5.6 em
		[Optimization-based, text width=8.5em
		[\cite{9744540},leaf,text width=6.6 em]
		]
		[Artificial Potential Field, text width=8.5em
		[\cite{yao2020path},leaf,text width=6.6 em]
		]
		[Dynamic Window, text width=8.5em
		[\cite{yang2022automatic},leaf,text width=6.6 em]
		]
		[Data-driven, text width=8.5em
		[\cite{liu2020mapper},leaf,text width=6.6 em]
		]
		]
		]
		[Motion Control,text width=9em
		[PID{,} Preview Tracking Control{,} Sliding Mode Control{,} MPC{,} MPPI,
		leaf,text width=24.3 em]
		]
		]
		[End-to-End AD, text width=5.7em
		[Imitation Learning,text width=9 em
		[Behavior Cloning{,}
		Inverse Reinforcement Learning~\cite{zhu2020off},
		leaf,text width=24.3 em]
		]
		[Reinforcement Learning,text width=9 em
		[Value-based{,} Policy-based{,}
		Actor-critic \cite{weerakoon2024vapor}{,} ...,
		leaf,text width=24.3 em]
		]
		]
		[AD Dataset, text width=5.7 em
		[Traversability Estimation,text width=9 em
		[ORFD~\cite{orfd}{,}
		M3-GMN~\cite{m3gmn}{,} ...,
		leaf,text width=24.3 em]
		]
		[Semantic Segmentation,text width=9 em
		[DeepScene~\cite{valada2017deep}{,} YCOR~\cite{maturana2018real}{,} ...,
		leaf,text width=24.3 em]
		]
		[Other Datasets,text width=9 em
		[OPEDD~\cite{neigel2020opedd}{,} TartanDrive~\cite{triest2022tartandrive}{,} ...,
		leaf,text width=24.3 em]
		]
		]
		]
		]
	\end{forest}
	\caption{Taxonomy of autonomous driving in unstructured environments.}
	\label{fig:taxonomy}
\end{figure*}
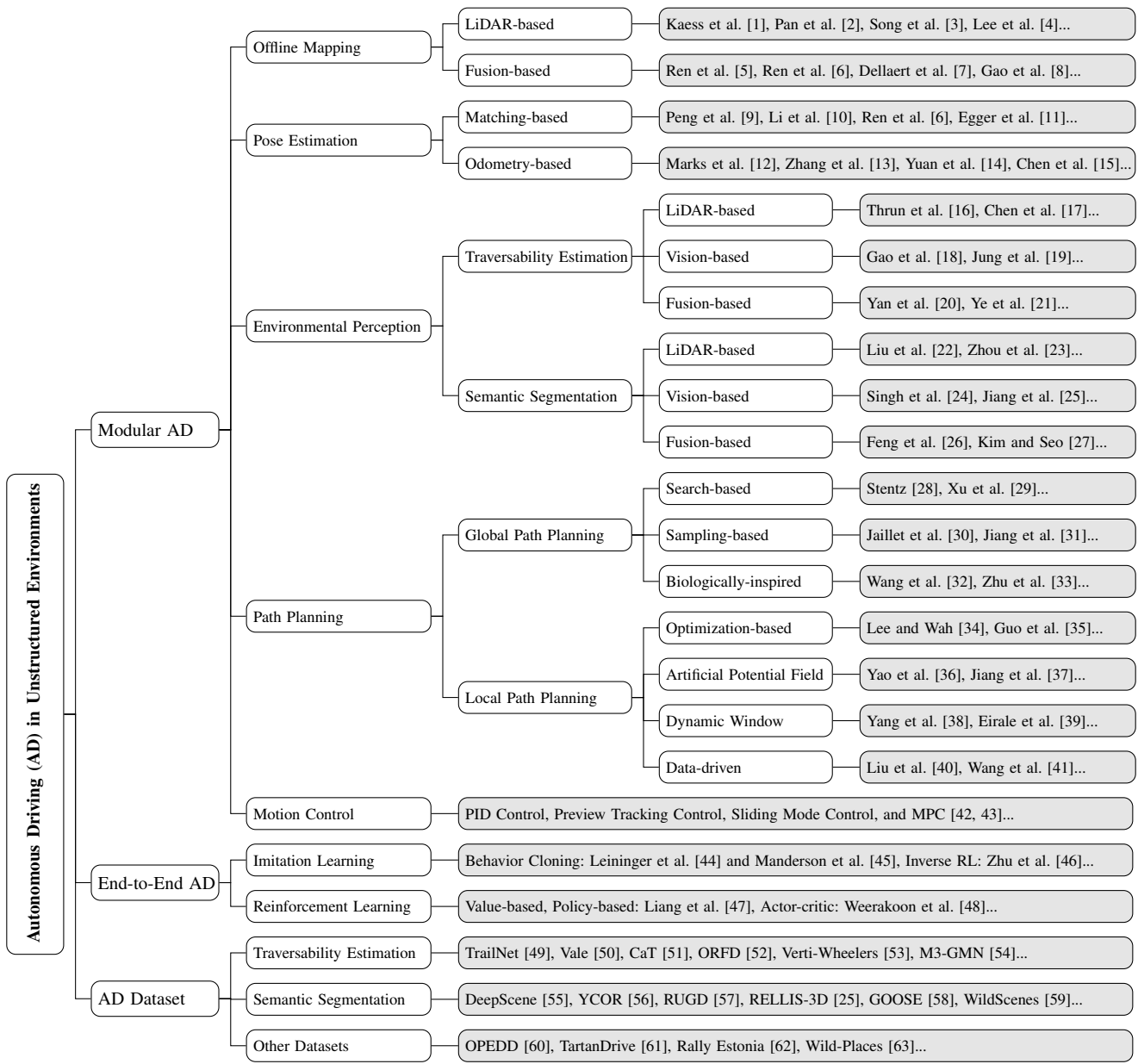
\begin{figure*} [t]
	\centering
	\includegraphics[width=0.98\textwidth]{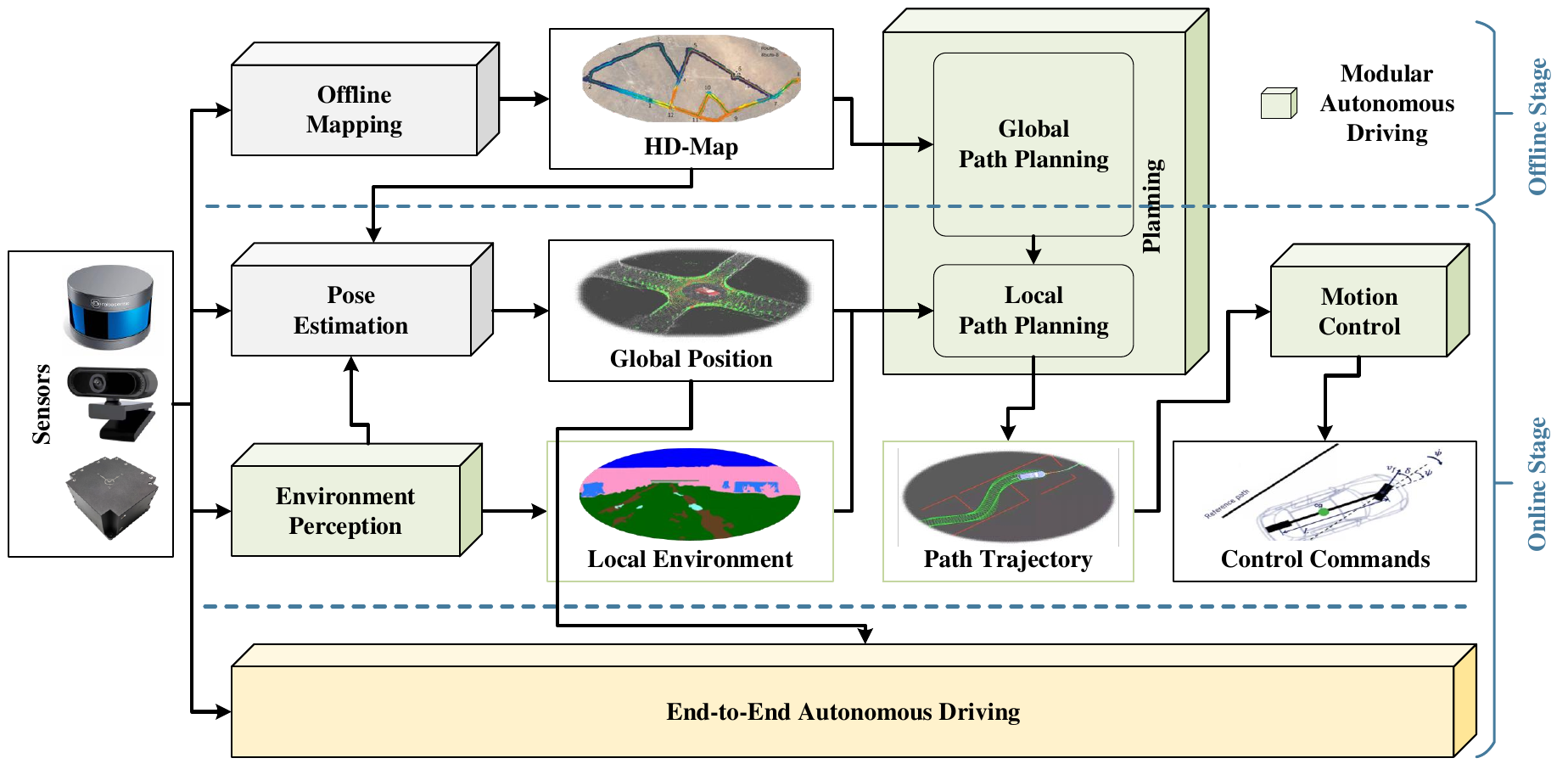}
	\caption{Typical components of the autonomous driving system. The modules for autonomous driving systems in both unstructured and structured environments are fundamentally similar, comprising offline mapping, pose estimation, environmental perception, path planning, and motion control.}
	\label{fig:system}
\end{figure*}

\begin{figure*} [t]
	\centering
	\includegraphics[width=0.98\textwidth]{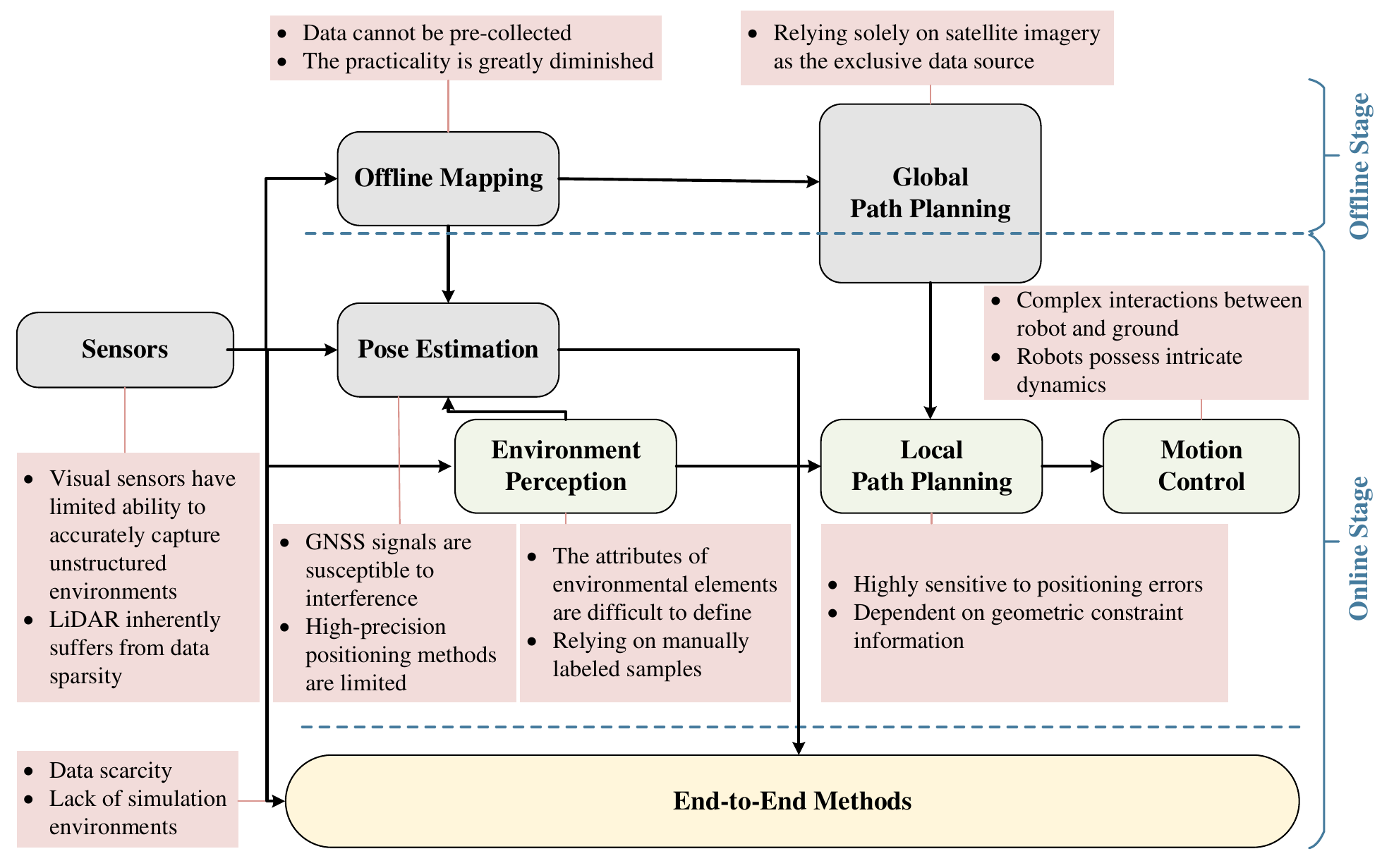}
	\caption{Challenges faced by autonomous driving systems in unstructured environments.}
	\label{fig:failure}
\end{figure*}

In our discussion, we define an unstructured environment as off-road, suburban, or rural areas with unstructured or minimally structured roads, where well-defined routes and typical driving cues, such as road signs and traffic signals, are lacking. Autonomous driving in unstructured settings can address the needs of diverse fields such as agriculture, construction and mining, logistics and delivery, rescue and exploration, and military operations. Compared to structured environments, these unstructured environments present entirely different environmental characteristics, posing more severe challenges to the autonomous driving technology of vehicles. Specifically, as illustrated in Fig.~\ref{fig:features}, there are six inherent characteristics of unstructured environments:

(1) \textbf{Diversity of Environmental Types}: Unstructured environments, such as grasslands, deserts, forests, and mountainous regions, exhibit significant differences in terrain, landforms, and features. Additionally, changing environmental factors like seasons and weather can further impact the geometric structure and appearance of unstructured environments.

(2) \textbf{Vast Variety of Elements}: The environmental features in unstructured environments are predominantly natural objects, which not only come in numerous types but also exhibit various forms. Some elements may even be challenging to distinguish manually.

(3) \textbf{Ambiguity of Semantic Categories}: In unstructured environments, many environmental features often exist in a tangled and mixed manner. The spatial relationships and semantic connections between these features are intricate, with fuzzy boundaries that make it exceedingly difficult to accurately define the semantic category to which a particular element belongs. Furthermore, even within the same semantic category, environmental features may display radically different geometric shapes and appearances.

(4) \textbf{Disorderly Scene Structure}: Unstructured environments lack clear traffic features such as lane markings and paved surfaces. The traversable areas are typically irregular in shape, with ambiguous orientations and variable structures, often lacking distinctly identifiable roads. Additionally, natural features in unstructured environments are distributed without any fixed pattern.

(5) \textbf{Complexity of Road Conditions}: The surface materials of roads in unstructured environments are diverse, including soil, gravel, grass, and mud. Furthermore, the roads are often rugged and uneven, with significant vertical fluctuations or lateral inclines. The geometric characteristics of roads in unstructured environments, including curvature and slope, are also much more complex and variable.

(6) \textbf{Vulnerability of Satellite Signals}: In complex unstructured environments such as canyons and forests, satellite signals are easily obstructed or attenuated. During transmission, signals may also reflect or scatter due to obstructions, leading to significant reductions in positioning accuracy. Additionally, electromagnetic interference commonly present in military application scenarios poses serious disruptions to satellite signals.

Through the above analysis, it is evident that the complexity of unstructured environments significantly exceeds that of structured environments. Highly disordered settings contain numerous environmental elements with varying shapes, fuzzy boundaries, and easily confused semantic categories. Moreover, the lack of effective prior information exacerbates the unknowns and uncertainties faced by autonomous driving systems in unstructured environments. These factors greatly increase the implementation difficulty of autonomous driving technology in such settings and impose stricter requirements on the intelligent levels of core aspects such as scene representation, environmental understanding, and behavioral decision-making.

Despite the challenges, autonomous driving in unstructured environments has attracted early research interest. For example, the 2004 DARPA Grand Challenge took place in a desert setting~\cite{ebadi2023present}. Some reviews~\cite{teji2023survey} have summarized research in this area.
\cite{guastella2020learning} focus on perception in autonomous driving within unstructured environments, while \cite{borges2022survey} and \cite{islam2022off} examine traversability estimation in such contexts. \cite{wijayathunga2023challenges} address both perception and path planning for autonomous driving in unstructured environments. The survey in~\cite{wang2024survey} is dedicated to path planning, and \cite{mortimer2024survey} emphasize datasets relevant to autonomous driving in unstructured environments.
However, existing reviews on autonomous driving in unstructured environments often focus on specific aspects such as traversability analysis, path planning, and datasets, while neglecting the holistic view of autonomous driving as an integrated system. Moreover, current reviews lack investigation into offline mapping and pose estimation for autonomous driving in unstructured environments. In this survey, we present a comprehensive review of autonomous driving in unstructured environments, as shown in Fig.~\ref{fig:taxonomy}.

The workflow of a traditional modular autonomous driving system can be summarized as shown in Fig.~\ref{fig:system}. During the offline preparation phase, the system first employs the offline mapping module to create a high-precision prior map of the target environment. Next, based on the actual autonomous task requirements of the vehicle, the global path planning module generates a global path to provide directional guidance for autonomous driving.
Once the online driving phase begins, the system receives real-time data collected by onboard sensors. The pose estimation module determines the instantaneous position and orientation of the vehicle, while the environmental perception module analyzes the real-time situation of the surrounding environment. The local path planning module then generates a local driving path that the vehicle should follow, and the motion control module converts these paths into specific control commands.

The close collaboration among these modules enables the autonomous driving functionality of the system. However, as illustrated in Fig.~\ref{fig:failure}, when facing complex and unknown unstructured environments, each core module of the autonomous driving system encounters varying degrees of challenges, significantly hindering the normal operation of the workflow.

Autonomous driving systems first feature a map construction module that creates detailed representations of the environment, storing static data to support pose estimation and planning. However, in unstructured environments like disaster zones, pre-collection of data is often impossible, limiting the map's usability~\cite{ren2021lidar,xue2019imu}. The mapping process generally utilizes SLAM technology alongside various sensor types, with LiDAR methods proving particularly effective~\cite{fu2020lidar,fu2020fast}. Even when data can be gathered, the absence of key elements like signs significantly reduces the maps' practical value, while natural changes in terrain require frequent updates, leading to increased operational costs. In this survey, we will review the challenges associated with offline mapping in unstructured environments, along with various methods, including LiDAR-based and fusion-based approaches. We will review methods for pose estimation in such settings, including matching-based and odometry-based approaches.

The goal of environmental perception in autonomous driving is to comprehend the vehicle's surroundings~\cite{wang2022sts}. In unstructured environments, the vehicle's primary focus is to identify obstacles and determine traversable areas. We first review traversability estimation algorithms, including LiDAR-based, vision-based, and fusion-based methods~\cite{xue2021lidar}. With the advancement of deep learning technologies~\cite{min2021attentional}, semantic segmentation methods have been applied to unstructured scenes, enhancing the vehicle's comprehensive understanding of its surroundings.

Path planning in unstructured environments presents several key challenges that differ from those in structured environments. These include managing high uncertainty and dynamic changes, navigating complex state spaces, dealing with limited prior knowledge, meeting real-time operation demands, and addressing the complexities of long-term planning~\cite{zuo2015hierarchical}. We will delve into these challenges and examine global and local path planning algorithms designed for unstructured environments, which can be categorized into the following approaches: search-based, sample-based, optimization-based, artificial potential field-based, dynamic window approach-based, biologically-inspired, and data-driven methods.

Vehicle motion control acts as the vehicle's ``hands and feet." Unique modeling challenges emerge when navigating complex unstructured terrains, resulting in intricate interactions between the vehicle and the ground, which complicate high-speed control. Unpredictable terrains, influenced by conditions like rain or snow, lead to traction variations and uneven load distribution that affect stability. Moreover, each vehicle's characteristics, such as weight distribution, play a crucial role in its response to inputs, highlighting the importance of understanding these dynamics for effective control. We will review motion control methods for unstructured environments.

The aforementioned modular autonomous driving systems offer strong interpretability and make it easier to pinpoint problem areas, but information can be lost during transmission~\cite{min2024driveworld}. As shown in Fig.~\ref{fig:system}, end-to-end autonomous driving uses a single network to directly output control signals from sensor data, significantly reducing information loss compared to modular autonomous driving algorithms~\cite{zhu2024sora}. Additionally, unstructured scenes are complex and variable, making data collection and labeling challenging. For instance, end-to-end autonomous driving based on imitation learning only requires the collection of driver data for training, facilitating rapid advancements in autonomous driving technology for unstructured environments. This paper introduces current end-to-end autonomous driving algorithms focusing on unstructured scenes, and this field is expected to develop rapidly in the future.

In recent years, deep learning has significantly advanced autonomous driving technology, heavily relying on well-annotated datasets~\cite{min2023occupancy}. Several datasets designed for unstructured environments, such as RELLIS-3D~\cite{rellis3d}, RUGD~\cite{rugd}, and ORFD~\cite{orfd}, have been introduced. These datasets primarily focus on traversability estimation and semantic segmentation tasks and are collected using sensors like RGB cameras, LiDAR, stereo cameras, and IMUs. While these datasets have propelled the development of autonomous driving technology in unstructured scenarios, they remain relatively small in scale compared to those for structured urban environments, indicating a need for further expansion and improvement.

This review aims to provide a comprehensive overview and analysis of the current state, challenges, and future research directions in autonomous driving technology for unstructured environments. It is designed to offer valuable insights for newcomers entering this field and to foster critical discussion among established researchers. In
Section~\ref{mapping} we pinpoint offline mapping in unstructured environments. In Section~\ref{pose} we review pose estimation methods,
followed by environmental perception methods in Section~\ref{perception}. Section~\ref{planning} is dedicated to path planning methods,
while motion control is discussed in Section~\ref{control}. End-to-end autonomous driving methods are
discussed in Section~\ref{e2e}. Section~\ref{data} reviews available datasets
related to autonomous driving in unstructured environments. In Section~\ref{future} we present several future research directions. Finally, in Section~\ref{conclusion}, we provide relevant conclusions about autonomous driving in unstructured environments. 

The main contributions of this review are:
\begin{itemize}
	\item A thorough examination of recent advancements in autonomous driving within unstructured environments, including both deep philosophical insights and detailed discussions.
	
	\item A detailed analysis of literature related to various aspects of autonomous driving in unstructured environments, such as offline mapping, pose estimation, environmental perception, path planning, motion control, end-to-end autonomous driving, and datasets.
	
	\item An evaluation of the existing challenges and limitations for autonomous driving in unstructured environments, along with an exploration of potential research directions to guide and stimulate future progress.
\end{itemize}
\section{Offline Mapping} 
\label{mapping}

In autonomous driving systems, the map-building module uses offline pre-collected data to create high-precision prior maps of the target environment. These maps effectively store and represent critical information about the static environmental features and road topology, providing valuable prior data for the subsequent pose estimation and global path planning modules. However, in complex and unknown unstructured environments, the usability of the offline mapping module is often difficult to ensure.

\subsection{Challenges of Offline Mapping in Unstructured Environments}

\begin{itemize} 
	
	\item \textbf{Inability to Pre-Collect Data}: In autonomous missions targeting unstructured environments, such as battlefield reconnaissance or disaster recovery in rubble-strewn areas, a common challenge is the unknown and restricted nature of these environments. This means that before the mission begins, the vehicle is typically unable to enter the target area to collect data, fundamentally limiting the usability of the map-building module.
	
	\item \textbf{Significantly Reduced Practicality}: Even in unstructured environments where pre-collection of data is possible, the lack of key traffic elements like signs, markings, and paved roads leads to a drastic reduction in the amount of useful information contained in the prior maps. This severely diminishes their practical value in autonomous driving systems. Additionally, natural terrain and features in unstructured environments can change significantly with the seasons and weather, requiring frequent data collection to update the prior maps. This significantly increases the operational and maintenance costs of the system.
	
\end{itemize}

The map construction process typically revolves around Simultaneous Localization and Mapping (SLAM) as the core technology, complemented by manual labeling and precise surveying techniques for both online and offline applications. Based on the types of sensors employed, the mapping methods for autonomous driving can be categorized into two main types: LiDAR-based methods and multi-sensor fusion methods, as shown in Table~\ref{tab:offmap}. Thanks to their exceptional ranging accuracy and robust performance under varying lighting conditions, LiDAR-based mapping methods have seen extensive application in the field of map construction.

\subsection{LiDAR-based Methods} 


\cite{wolcott2017robust} construct a pose graph incorporating odometry, laser scan matching, and GPS prior constraints, solving the nonlinear least squares problem using the iSAM~\cite{kaess2008isam} method to optimize robot poses. Point clouds are reprojected into a global frame and accumulated into a global point cloud, stored via sparse histograms. The expectation-maximization algorithm is then applied to each histogram column to generate Gaussian mixture models for each grid cell, capturing z-height and reflectivity distributions, thus creating offline maps for online localization. \cite{fu2020lidar} propose a two-stage fusion algorithm for LiDAR scan matching in off-road environments, combining correlative scan matching (CSM) for large offsets and local scan matching techniques for smaller offsets, significantly enhancing accuracy and robustness in complex terrains. 

Towards the large-scale unstructured environments, a robust off-line mapping system~\cite{ren2021towards,ren2021lidar} integrating GNSS and LiDAR data using a factor graph framework~\cite{dellaert2017factor}, introducing a new degeneration indicator for point cloud registration to maintain precision despite GNSS outages.
\cite{li2020localization} utilize LiDAR and Global Navigation Satellite System (GNSS) during the offline mapping process to record tree trunk positions and collect point cloud data of the forest environment. The data is then processed to extract trunk points, project them onto a 2D plane, and perform Delaunay triangulation to create a global Delaunay Triangulation graph. This ensures an accurate global forest map for real-time localization of autonomous robots. For low-speed vehicles in unstructured environments, a mapping system~\cite{gao2021fully} combining multiple sensors into a factor graph achieves globally consistent maps through robust two-stage optimization, correcting RTK measurements and detecting LiDAR odometry degeneration.
\begin{table*}[htbp] 
	\caption{Offline mapping methods for unstructured environments.} 
	\begin{center}
		\setlength{\tabcolsep}{0.2mm}{
			\begin{tabular}{c|c|c|c}
				\hline
				\bf Reference & \bf Task & \bf Scenario & \bf Sensors  \\ 
				\midrule
				
				\cite{wolcott2017robust} & LiDAR Mapping & Multi-scene & GPS/LiDAR \\ 
				\cite{fu2020lidar} & Point Clouds Registration & Off-road & LiDAR \\
				\cite{pan2020gem} & Dense Elevation Mapping & KITTI/Campus & LiDAR/Camera \\ 
				\cite{chahine2021mapping} & Mapping & Snowy Forest & GPS/IMU/LiDAR/Camera \\ 
				\cite{ren2021towards,ren2021lidar} & LiDAR Mapping & Off-road & GNSS/LiDAR \\ 
				\cite{gao2021fully} & LiDAR Mapping &  Residential Area & RTK/IMU/LiDAR \\ 
				\cite{cui2023real} & LiDAR Mapping & Featureless Area & RTK/IMU/LiDAR \\ 
				\cite{song2023safety} & Semantic Mapping & Off-road & GPS/IMU/LiDAR \\ 
				\cite{lee2024three} & Online Mapping & Unpaved Road & GNSS/LiDAR/Camera \\ 
				\cite{zou2024lta} & LiDAR Mapping & Multi-scene & IMU/LiDAR \\
				\cite{lim2024outlier} & LiDAR Mapping & Multi-scene & LiDAR \\
				\cite{hu2024ms} & LiDAR Mapping & Multi-scene & LiDAR \\ 
				
				\hline
			\end{tabular}
			\label{tab:offmap}
		}
	\end{center}
\end{table*}

Further advancements involve real-time 3D point cloud mapping in featureless environments~\cite{cui2023real}, utilizing Real Time Kinematic (RTK)  and IMU data with Normal Distributions Transform (NDT) for height calibration, demonstrating robust mapping performance in real-world excavator scenes. The creation of safety-assured semantic maps for autonomous engineering vehicles~\cite{song2023safety} integrates hazardous obstacle detection and semantic information, enhancing safety in unstructured terrains. 

Long-term association LiDAR-IMU odometry systems, like LTA-OM~\cite{zou2024lta}, use corrected historical maps to provide global constraints during mapping, reducing drift and improving consistency over multiple sessions. 
\cite{lim2024outlier} presents the outlier-robust mapping techniques that leverage fast ground segmentation and graduated non-convexity (GNC) for handling significant data outliers, ensuring reliable mapping in dynamic environments. Finally, MS-Mapping~\cite{hu2024ms} addresses data redundancy and pose graph scalability using Gaussian mixture model and Wasserstein distance-based keyframe selection, achieving high efficiency and accuracy in large-scale environments. 

\subsection{Fusion-based Methods}
The GEM system~\cite{pan2020gem} offers a globally consistent elevation mapping framework, representing global maps as deformable submaps to maintain consistency during trajectory corrections, supported by GPU-CPU coordinated processing for real-time performance. In snowy forest environments, a sensor fusion framework~\cite{chahine2021mapping} utilizes cameras, LiDARs, and other sensors to reconstruct natural settings, incorporating innovative visual map registration and ICP-inferred loop closure to address position and attitude drift, achieving high map reconstruction quality. LiDAR-visual odometry methods for unpaved roads~\cite{lee2024three} improve mapping accuracy by integrating LiDAR and visual data, employing novel interpolation techniques and intensity-weighted motion estimation to mitigate dust effects. 

\section{Pose Estimation}
\label{pose}
In autonomous driving systems, the pose estimation module uses real-time sensor data and known prior information to estimate the position and orientation of the vehicle, providing essential location data for the subsequent planning and decision-making modules. However, when operating in complex and unknown unstructured environments, the pose estimation module often faces significant challenges in delivering high-precision global positioning.

\subsection{Challenges of Pose Estimation in Unstructured Environments}

\begin{itemize} 
	
	\item \textbf{GNSS Signal Susceptibility to Interference}: In complex unstructured environments, GNSS signals are highly prone to interference from obstacles such as mountains, trees, and rocks, causing reflection or scattering and triggering multipath effects, which drastically reduce positioning accuracy. Furthermore, the presence of electromagnetic interference poses a serious threat to GNSS availability. Such interference can distort or completely block GNSS signals, severely affecting the accuracy and stability of pose estimation results.
	
	\item \textbf{Limited High-Precision Positioning Methods}: In unknown unstructured environments, the lack of high-precision prior maps renders map-matching-based positioning methods unusable. While dead reckoning-based methods can be employed to some extent, their inherent issue of cumulative errors remains difficult to resolve, and the complexity of unstructured environments can further exacerbate this error accumulation. As a result, high-precision positioning options in the pose estimation module are extremely limited, making it challenging to achieve robust, high-accuracy localization.
	
\end{itemize}

Currently, pose estimation methods used in autonomous driving can be categorized into three main types: those based on GNSS, those based on map-matching, and those based on odometry. 
Pose estimation methods based on GNSS can be adversely affected by satellite obstruction or multipath propagation effects, leading to significant drift in the output pose estimation results and resulting in positioning failures.
Pose estimation methods based on map matching achieve accurate estimation of the ground unmanned platform's pose by aligning real-time sensor observations with a pre-constructed high-precision map. This process primarily involves two key steps: observation matching and optimization correction.
Pose estimation methods based on odometry take the initial position of the ground unmanned platform as the origin and use sensor data obtained online to iteratively calculate the changes in pose relative to the previous time point.

\subsection{Matching-based Methods} 

\begin{table*}[htbp]
	\caption{Pose estimation methods using prior HD-Map for unstructured environments.}
	\begin{center}
		\setlength{\tabcolsep}{1.0mm}{
			\begin{tabular}{c|c|c|c}
				\midrule
				\bf Reference & \bf Prior Map Type & \bf Scenario & \bf Sensors  \\ 
				\midrule
				
				\cite{wolcott2017robust} & \multirow{7}{*}{LiDAR Point Cloud Map} & Multi-scene & IMU/LiDAR \\ 
				\cite{li2020localization} & & Forest & LiDAR \\ 
				\cite{ren2021lidar} & & Off-road & IMU/LiDAR \\ 
				\cite{egger2018posemap} & & Multi-scene & IMU/LiDAR \\ 
				\cite{zou2024lta} & & Multi-scene & IMU/LiDAR \\ 
				\cite{peng2022roll} & & Changing Scene & IMU/LiDAR \\ 
				\cite{hroob2023learned} & & Changing Scene & IMU/LiDAR \\ 
				\midrule
			\end{tabular}
			\label{tab:loc-hd}
		}
	\end{center}
\end{table*}

\subsubsection{Matching with High-Definition Map} 

For autonomous driving in unstructured environments, map-matching-based pose estimation method with High-Definition Map (HD-Map) is crucial for precise and safe navigation, as shown in Table~\ref{tab:loc-hd} 
Some methods focus on the robustness and efficiency of localization within prior maps. \cite{wolcott2017robust} introduce a method utilizing multiresolution Gaussian mixture models (GMMs) to create maps capturing both three-dimensional structure and ground-plane reflectivity. This method allows for fast and accurate multiresolution inference, making real-time optimal localization feasible even under adverse conditions like heavy snowfall. The use of GPU processing further accelerates computation, enabling the system to handle all measured 3D points without spatial downsampling, thereby maintaining high localization accuracy. 

Similarly, \cite{li2020localization} propose using Delaunay triangulation to match local point clouds to a global tree map for autonomous robots in forest environments. This approach achieves fast and real-time global localization, leveraging the Delaunay triangulation to solve vehicle tracking problems solely with LiDAR data, proving effective in complex forest environments. Additionally, \cite{ren2021lidar} present a robust 3D offline mapping and online localization approach for off-road scenarios. The method includes a novel degeneracy indicator integrated into a factor graph framework, which helps detect and handle the degeneration state of scan matching algorithms. The system can seamlessly switch between map matching and LiDAR odometry modes, providing high accuracy and flexible localization solutions in challenging off-road conditions. 

Another significant focus is on long-term localization strategies that adapt to changing environments. The PoseMap~\cite{egger2018posemap} addresses the challenges of reliable localization over extended periods by extracting distinctive features from range measurements and bundling them into local views with observation poses. This method maintains a distributed map composed of submaps, offering online map extension capabilities and a lifelong localization strategy robust to environmental changes such as moving vehicles and vegetation. Similarly, the LTA-OM system~\cite{zou2024lta} supports lifelong mapping by allowing users to store results from current sessions for future use, including corrected map points and optimized odometry. This system can load previously generated maps as prior maps for current sessions, ensuring consistent localization and mapping across multiple sessions, particularly beneficial for long-term deployments of autonomous robots in dynamic environments, eliminating the need to rebuild maps for each new task.

ROLL~\cite{peng2022roll} incorporates temporary mapping to address long-term scene changes, and this system activates mapping processes when matching quality is low, discarding unreliable global matching poses and merging temporary key-frames into the global map once reliable association is achieved. Moreover, \cite{hroob2023learned} introduce a stability scan filter that learns and infers the motion status of objects, identifying stable landmarks for robust localization. Using an unsupervised labeling of LiDAR frames and a regression network, it effectively identifies long-term stable objects in agricultural environments, significantly improving localization performance by focusing on stable landmarks despite continuous environmental changes. 

\subsubsection{Matching with Lightweight Map} 

\begin{table}[htbp] 
	\caption{Pose estimation methods using lightweight maps for unstructured environments.} 
	\begin{center}
		\resizebox{0.7\textwidth}{!}{
			\begin{tabular}{c|c|c}
				\midrule
				\bf Reference & \bf Map Type & \bf Sensors \\ 
				\midrule
				
				\cite{burger2019unstructured} & Road Network & LiDAR \\ 
				\cite{ort2019maplite} & Topometric Map & LiDAR \\
				\cite{omama2023alt} & Topometric Map & LiDAR/Camera \\
				\cite{si2022tom} & Topometric Map & Camera \\
				\cite{ankenbauer2023global} & Semantic Object Map & LiDAR/Camera \\ 
				
				\midrule
			\end{tabular}
			\label{tab:loc-light}
		}
	\end{center}
\end{table}

In unstructured environments, while obtaining and maintaining HD-Map is costly, certain lightweight maps, such as topometric maps, are more cost-effective. These maps serve as efficient representations of the environment and can provide reliable and stable global constraints for the navigation and localization of autonomous vehicles~\cite{si2022tom}, as shown in Table~\ref{tab:loc-light}. 

One innovative method is presented by~\cite{burger2019unstructured}, which combines filter-based road course tracking and SLAM in unstructured rural areas lacking road markings or HD-Map. This approach employs a novel B-Spline measurement model tracked with an unscented Kalman filter, enhancing vehicle localization by correcting global position offsets through iterative closest point algorithms. \cite{ort2019maplite} introduce MapLite, a system designed for autonomous intersection navigation without detailed prior maps or GPS localization. MapLite uses a sparse topometric map from OpenStreetMap~\cite{haklay2008openstreetmap} and integrates onboard sensors to segment the road, register the topometric map to sensor data, and plan trajectories using a variational path planner. This method successfully navigates over 15 km of rural roads with more than 100 intersection traversals, showcasing its effectiveness in unstructured environments. 

Based on~\cite{ort2019maplite}, \cite{omama2023alt} introduce ALT-Pilot, a system that utilizes publicly available road network information and crowdsourced language landmarks for navigation. ALT-Pilot employs a probabilistic multimodal localization algorithm integrating LiDAR, image, and language data, achieving high localization accuracy and robustness in both simulations and real-world scenarios. Similarly, \cite{si2022tom} propose TOM-odometry, a generalized localization framework that integrates odometry and topometric maps for GPS-denied scenarios. This framework employs odometry as a kinematic model for relative positioning, while the topometric map serves as a virtual global sensor to correct drift errors.

In~\cite{ankenbauer2023global}, lightweight semantic object maps are incorporated to guide relocalization in unstructured environments. This method associates and registers the vehicle's local semantic object map with a compact semantic reference map built from various viewpoints, time periods, and modalities. A graph-based data association algorithm ensures robustness to noise, outliers, and missing objects, significantly improving global localization accuracy in non-urban and urban settings. 

\subsubsection{Place Recognition / Relocalization} 

\begin{table*}[htbp] 
	\caption{Place recognition methods for unstructured environments.} 
	\begin{center}
		\setlength{\tabcolsep}{1.0mm}{
			\begin{tabular}{c|c|c|c}
				\midrule
				\bf Reference & \bf Environment Encoding \bf Format & \bf Scenario & \bf Sensors  \\ 
				\midrule
				
				\cite{tian2022stv} & Scan Context & Multi-scene & LiDAR \\ 
				\cite{ou2023place} & Spatial Binary Pattern & Orchard & LiDAR \\ 
				\cite{yuan2024btc} & Binary and Triangle Combined Descriptor & Multi-scene & LiDAR \\ 
				\cite{ramezani2023deep} & Global and Local Descriptor & Natural Scene & LiDAR/Camera \\ 
				\cite{oh2024evaluation} & Diverse Descriptors for Evaluation & Dense Forest & IMU/LiDAR \\ 
				\cite{barros2024pointnetpgap} & PointNetPGAP-based Descriptor & Horticulture & LiDAR \\ 
				
				\midrule
			\end{tabular}
			\label{tab:pr}
		}
	\end{center}
\end{table*}

Place Recognition (PR) technology enables vehicles to identify previously visited environments, playing a crucial role in autonomous driving. This technology can be utilized to reduce drift errors of SLAM systems, ensuring reliable localization and globally consistent map construction. In unstructured environments, the PR task is particularly challenging due to the dynamic nature and sparse, diverse features of the surroundings. Such conditions often render methods~\cite{kim2018scan,kim2021scan,zhang2024osk} designed for urban environments ineffective. Currently, there are some studies that focus on PR technology in unstructured environments, as shown in Table~\ref{tab:pr}. 

\cite{tian2022stv} propose STV-SC, which enhances the scan context \cite{kim2018scan} algorithm by incorporating segmentation and temporal verification. This method includes a range image-based 3D point segmentation algorithm to extract key structured information and a three-stage process for loop detection, significantly improving recall rates compared to state-of-the-art algorithms. \cite{ou2023place} introduce a novel LiDAR-based PR algorithm that utilizes the spatial binary pattern to encode 3D spatial information into an eight-bit binary pattern. The two-stage hierarchical re-identification process with attention score maps further enhances the method’s effectiveness in orchard and public datasets. Another innovative method is~\cite{yuan2024btc}, which combines binary and triangle descriptors to achieve robust 3D place recognition. This method forms triangles from keypoints and uses their side lengths to create a global descriptor invariant to pose changes, significantly improving precision and adaptability, especially in scenarios with large viewpoint variations. 

Further advancements in this domain involve integrating deep learning techniques for improved robustness in challenging scenarios. \cite{ramezani2023deep} present a method that integrates LiDAR and image data through a self-supervised 2D-3D feature matching process, leveraging deep learning for LiDAR feature extraction and relative pose estimation between point clouds. This approach improves robustness and accuracy in multi-robot relocalization tasks in unstructured environments. \cite{oh2024evaluation} evaluate handcrafted and learning-based models in dense forest environments, integrating the Logg3dNet model into a robust 6-DoF pose estimation system. 

\cite{barros2024pointnetpgap} introduce PointNetPGAP-SLC, a model combining a global average pooling aggregator and a pairwise feature interaction aggregator to enhance descriptor generation. The Segment-Level Consistency (SLC) model used during training improves the robustness of descriptors. Evaluations on the dataset demonstrate that PointNetPGAP improves PR performance in horticultural settings by addressing descriptor ambiguity and leveraging the SLC model for enhanced training. 

Moreover, the development of specific datasets and evaluation frameworks has been crucial in advancing PR techniques in natural environments. \cite{hausler2024towards} address the lack of benchmarks for natural, unstructured environments by introducing the Wild-Places~\cite{wideplaces} and WildScenes~\cite{wildscenes} datasets. These datasets enable comprehensive evaluation and development of localization and mapping algorithms in challenging unstructured settings, bridging the gap between urban and natural environments. 

\subsection{Odometry-based Methods} 

In autonomous driving, achieving accurate and reliable localization in unstructured environments without the aid of pre-existing maps is a significant challenge. The lack of maps can arise due to various reasons, such as the high cost and effort associated with creating and maintaining these maps, frequent changes in the environment, or the vehicle operating in previously unmapped areas. Such environments require real-time adaptation and robust algorithms to handle dynamic and complex terrains. Leveraging different sensor modalities and adaptive algorithms to improve localization accuracy and robustness is essential in these map-free scenarios. 


\begin{figure*}[htbp]
	
	\centering
	\includegraphics[width=16cm]{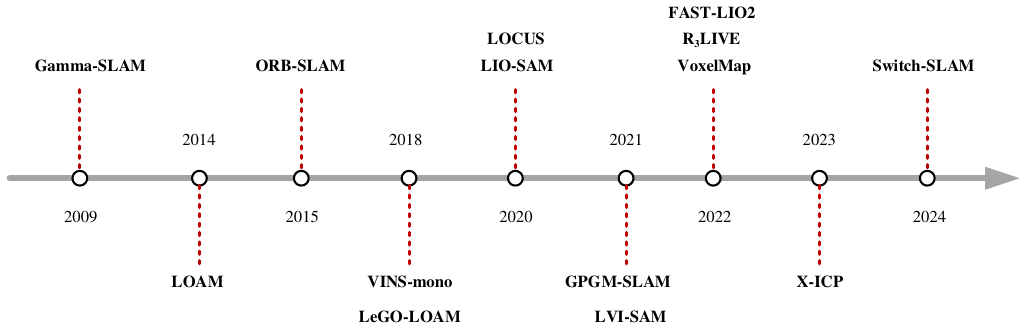} 
	\caption{SLAM-based pose estimation methods towards unstructured environments.} 
	\label{fig:slam_based_localization}
\end{figure*} 

In recent years, odometry and SLAM technologies, represented by vision and LiDAR~\cite{qin2018vins,mur2015orb,zhang2014loam,shan2018lego,shan2020lio,xu2022fast,lin2022r}, have gained widespread attention and have made significant advancements in both theory and application. These methods also offer solutions for map-less navigation and localization in unstructured environments. However, the characteristics of unstructured environments can easily lead to degradation or performance decline in such methods, making it one of the biggest challenges to ensure the stability and robustness of the algorithms. Therefore, some research efforts are directed towards these scenarios, as shown in Fig.~\ref{fig:slam_based_localization}. 

\cite{marks2009gamma} present Gamma-SLAM, an online stereo visual SLAM algorithm that uses a Rao-Blackwellized particle filter to estimate joint posterior distributions over poses and maps. This method incorporates visual odometry for accurate proposal distributions and represents maps through a Cartesian grid, maintaining a posterior distribution over elevation variance in each cell to effectively capture terrain features in unstructured environments. \cite{zhang2014loam} introduce LOAM, a real-time LiDAR odometry and mapping method that divides the problem into high-frequency odometry and low-frequency mapping algorithms. This approach ensures low drift and low computational complexity, making it suitable for real-time applications. Extending this concept, \cite{zhang2015visual} combine visual and LiDAR odometry (V-LOAM) to enhance localization and mapping by integrating high-frequency visual odometry for initial motion estimation and low-frequency LiDAR odometry for refinement, achieving robust performance even under aggressive motion and poor visual conditions.

Further advancements in this domain include methods that focus on optimizing computational efficiency and robustness in varying terrains. \cite{shan2018lego} introduce LeGO-LOAM, a lightweight and ground-optimized LiDAR odometry and mapping method that achieves real-time six degree-of-freedom pose estimation using low-power embedded systems. This method leverages the ground plane for segmentation and optimization, enhancing accuracy and computational efficiency. Based on~\cite{shan2018lego,wen2022agpc}, \cite{si2024rose} establish fixed and generative road surface models to constrain the vertical drift errors of autonomous vehicles in outdoor scenarios. Since ground point clouds can be continuously and stably obtained, this ground-constrained approach could also be suitable for feature-sparse off-road environments. 

\cite{giubilato2021gpgm} present GPGM-SLAM, which leverages Gaussian process gradient maps to improve place recognition robustness and facilitate loop closure detection in planetary analog environments, demonstrating superior performance compared to traditional visual SLAM methods. Additionally, \cite{chen2022direct} propose a lightweight and efficient LiDAR odometry method that handles dense point clouds without significant preprocessing, using an adaptive keyframing system and a custom iterative closest point solver to reduce computational overhead.

\cite{yuan2022efficient} introduce a novel voxel mapping method designed for LiDAR odometry, incorporating uncertainties from LiDAR measurement noise and pose estimation errors, and demonstrating superior accuracy and efficiency in various environments. \cite{wang2022lidar} present a LiDAR-based SLAM algorithm that leverages neighborhood information constraints to enhance feature point discrimination and prevent mis-registration in rugged terrains, showing significant improvements in localization and mapping accuracy. \cite{tuna2023x} introduce X-ICP, which incorporates localizability-aware constraints in LiDAR registration to ensure reliable pose updates in geometrically uninformative environments. \cite{dai2023intensity} propose an intensity-enhanced LiDAR SLAM framework that leverages both geometric and intensity information from LiDAR data to improve localization accuracy in sparse feature environments. 

The strategy of multi-sensor fusion has gained significant attention in recent years as an important means to effectively enhance the reliability and robustness of SLAM technology. For instance, \cite{xue2019imu} integrate IMU, wheel encoder, and LiDAR odometry to estimate vehicle motion, using IMU data to correct LiDAR scan distortions and provide better initial guesses for LiDAR odometry. This method demonstrates improved accuracy and high-frequency localization in various environments without prior information. Similarly, \cite{zhang2018laser} present a modularized data processing pipeline that integrates range, vision, and inertial sensing to estimate motion and construct maps through multilayer optimization, ensuring high accuracy and low drift even under aggressive motion and environmental changes. 

\cite{palieri2020locus} present LOCUS, a multi-sensor LiDAR-centric solution that integrates multiple sensing modalities and employs a robust scan-matching algorithm to ensure continuous operation despite sensor failures. This system demonstrates superior localization accuracy and robustness, particularly in challenging settings such as the DARPA Subterranean Challenge~\cite{ebadi2023present}. Additionally, \cite{shan2021lvi} introduce LVI-SAM, a tightly-coupled LiDAR-visual-inertial odometry framework that leverages a factor graph for global optimization. This method enhances visual features with LiDAR depth information and employs a robust failure detection mechanism, achieving high accuracy and robustness in GPS-denied and feature-sparse environments. 

\cite{lee2024switch} present Switch-SLAM, which uses a switching structure to transition between LiDAR and visual odometry based on detected degeneracies, ensuring robust and accurate localization across various challenging environments without the need for heuristic tuning. Additionally, \cite{lan2024highly} introduce a multi-sensor fusion localization system that integrates LiDAR, vision, and IMU data to achieve high accuracy and robustness in GNSS-denied environments, dynamically adjusting fusion weights in response to sensor degradation and demonstrating superior performance in urban areas, tunnels, and unstructured mountainous roads.  

Future research should investigate how mapping quality, including point cloud density and update frequency, affects pose estimation accuracy in unstructured terrains. For example, sparse LiDAR maps in low-feature environments such as deserts can lead to odometry drift and degraded localization, highlighting the need for adaptive strategies to mitigate map-induced errors.
\section{Environmental Perception}
\label{perception}

In autonomous driving systems, the environmental perception module analyzes the real-time situation of the surrounding environment using observational data collected by sensors. This analysis provides crucial information for the subsequent planning and decision-making modules. However, when navigating complex and unknown unstructured environments, the perception module faces numerous challenges.

\subsection{Challenges of Environmental Perception in Unstructured Environment}

\begin{itemize} 
	
	\item \textbf{Difficulty in Defining Environmental Features}: In unstructured environments, environmental features are hard to effectively define using manually preset rules or prior assumptions, which poses significant challenges to their recognition and understanding. For instance, navigable areas in unstructured environments may take various forms such as soil, grass, or gravel paths, often accompanied by mixed material composition and unclear boundaries. This makes rule-based methods for identifying navigable areas inadequate. Similarly, the classification of obstacles in unstructured environments is highly complex and ambiguous. The wide variety of potential obstacles, each with different shapes and characteristics, makes it impractical to set uniform detection standards manually. Furthermore, the properties of certain environmental features are closely tied to the physical characteristics of the vehicle itself, lacking absolutes, which adds another layer of complexity to defining environmental features.
	
	\item \textbf{Reliance on Manually Labeled Samples}: Despite the significant advancements in deep learning that provide powerful tools for the environmental perception module, their performance heavily depends on large quantities of manually labeled training samples. However, publicly available datasets for unstructured environments are still extremely scarce, forcing researchers to invest considerable time and resources into building datasets themselves. Additionally, due to the vast variety of features and the diversity of environmental types in unstructured environments, existing labeled data is often difficult to reuse effectively across different environments. As a result, when faced with new unstructured environments, extensive manual labeling efforts are required, leading to high data annotation costs. To complicate matters further, the ambiguity of semantic categories commonly found in unstructured environments causes confusion in categorizing environmental features and makes it difficult to distinguish feature boundaries. This further increases the difficulty and error rate of manual labeling.
	
\end{itemize}

In the environmental perception module, identifying passable areas is a core function to ensure the safety of autonomous driving. Additionally, another critical task within the environmental perception module is to segment key environmental factors such as trees, vehicles, and roads.
\subsection{LiDAR-based Methods}

\subsubsection{Traversability Estimation}

Terrain traversability analysis which focuses more on identifying suitable paths for unmanned vehicles, is crucial for achieving robotic autonomy in unstructured environments. Tasks closely linked to this process include road detection, ground segmentation, free-space detection, etc~\cite{xue2023traversability}. In unstructured environments, many methods designed for urban streets are inadequate because there are no pavements or lane markings, no curbs, or other artificial boundaries to distinguish road from non-road areas. Instead, the terrain is composed of natural objects with complex visual and geometric properties. Based on the sensors used, traversability estimation algorithms are classified into LiDAR-based, vision-based, and fusion-based.

Traversability estimation methods based on geometric information focus on extracting and analyzing geometric features from LiDAR point cloud to classify the passability of the environment. These methods can be further categorized into feature-based methods and model-based methods, as shown in Table~\ref{tab:lidar-ts}.

\paragraph{Feature-based Methods}

Feature-based methods for analyzing traversability primarily rely on extracting characteristics from LiDAR point cloud to assess terrain passability. A notable example is the Min-Max elevation mapping technique introduced by \cite{thrun2006stanley}. This approach begins by projecting the 3D LiDAR point cloud into a 2D grid map. The method evaluates the traversability of each grid cell by examining the height variations of points within that cell. Besides height differences, other parameters such as average height~\cite{douillard2010hybrid}, covariance~\cite{douillard2010hybrid}, slope~\cite{meng2018terrain}, and roughness\cite{neuhaus2009terrain} can also be employed to analyze traversability. These techniques are straightforward to implement and have relatively low computational demands. However, they often lack robustness and can struggle in adapting to environments characterized by rough terrain.

To enhance robustness in traversability analysis, various machine learning algorithms can be employed. In~\cite{lalonde2006natural}, the Expectation Maximization (EM) algorithm is utilized to fit a Gaussian Mixture Model (GMM) based on labeled statistical features, categorizing terrains in forest environments into three distinct groups. A Support Vector Machine (SVM) classifier, as described in~\cite{ahtiainen2017normal}, distinguishes between traversable and non-traversable regions, with each grid cell represented in an NDT traversability map (NDT-TM). In unstructured environments, a random forest classifier has been developed in~\cite{suger2016terrain} to identify traversable areas. Furthermore, \cite{lee2021self} detail the training of a Multi-Layer Perceptron (MLP) model for traversability mapping using a self-training algorithm. More recently, \cite{guan2021tns} combined geometric features, such as slope and step height, with semantic attributes to enhance traversability analysis.

With the rise of deep learning, many researchers have focused on applying it for feature extraction and traversability analysis. To effectively leverage Convolutional Neural Networks (CNNs), two primary strategies have been proposed for processing unordered 3D point clouds. The first strategy involves re-encoding point clouds to create a dense input tensor suitable for CNNs. For instance, some studies have transformed point clouds into elevation maps~\cite{chavez2018learning}, 2D Birds-Eye-View (BEV) maps~\cite{shaban2022semantic}, or range images~\cite{velas2018cnn}, enabling CNNs to detect traversable regions in an end-to-end manner. The second strategy employs architectures like PointNet~\cite{pointnet}, designed specifically for unordered 3D point sets. An end-to-end model GndNet~\cite{paigwar2020gndnet} integrates PointNet with a pillar feature encoding network for effective ground segmentation. Additionally, some approaches aim to predict traversability directly through CNNs; for example, \cite{seo2023scate} introduce a scalable framework for learning traversability from 3D LiDAR point clouds in a self-supervised fashion. Despite the promising results these learning-based methods have shown on various datasets, they often require substantial annotated datasets for training and struggle to adapt to different environments or LiDAR configurations, which poses limitations on their practical use.

\paragraph{Model-based Methods}
\begin{table}[htbp]
	\caption{LiDAR-based traversability estimation methods for unstructured environments.}
	\begin{center}
		\begin{tabular}{c|c|c}
				\hline
				\bf Type &\bf Reference & \bf Description \\ 
				\hline
				\multirow{8}{*}{Feature-based} &\cite{thrun2006stanley} &Min-Max Elevation\\
				&\cite{lalonde2006natural}&EM\\
				&\cite{ahtiainen2017normal}&SVM \\
				&\cite{suger2016terrain} &Random Forest Classifier\\
				&\cite{lee2021self} &MLP\\
				&\cite{guan2021tns} &Geometric Features\\
				&\cite{paigwar2020gndnet} &PointNet\\
				&\cite{seo2023scate} &Vehicle-terrain Interaction\\
				\midrule
				\multirow{10}{*}{Model-based} &\cite{chen2014gaussian}& 2D GPR\\
				&\cite{guizilini2020variational} & VHR\\
				&\cite{shan2018bayesian} & BGK\\
				&\cite{zhang2015ground} & MRF\\
				&\cite{rummelhard2017ground} & CRF\\
				&\cite{huang2021fast} & MRF\\
				&\cite{forkel2021probabilistic} &Recursive Gaussian State\\
				&\cite{rodrigues2020b}&B-spline \\
				&\cite{cai2023probabilistic} &Probabilistic\\
				&\cite{fan2021step} &2.5D Traversability\\
				&\cite{xue2023traversability} &BGK\\
				&\cite{xue2023contrastive} &Contrastive Label Disambiguation\\
				
				\hline
			\end{tabular}
		\label{tab:lidar-ts}
		\end{center}
\end{table}
In model-based approaches, the first step involves constructing a terrain model that reflects the continuity characteristics of the local environment, followed by an assessment of traversability based on this model. A widely used method in this domain is Bayesian non-parametric inference, which relies on the spatial correlation of point clouds to estimate a continuous terrain surface from observed data points. Among these techniques, Gaussian Process Regression (GPR) is particularly favored for traversability analysis tasks. However, executing GPR directly in a 2D space~\cite{vasudevan2009gaussian,lang2007adaptive} can be computationally intensive, often failing to meet the real-time requirements of unmanned ground vehicles. 

To address this challenge, \cite{chen2014gaussian} propose a method that transforms 2D GPR into a series of one-dimensional regressions, thereby reducing computational complexity. Nonetheless, this simplification tends to compromise the accuracy of the generated terrain model. An alternative approach, introduced in~\cite{guizilini2020variational}, utilizes Variational Hilbert Regression (VHR) for traversability analysis by projecting the point cloud into a Reproducing Kernel Hilbert Space (RKHS) for an approximate estimation of the terrain. 

More recently, Bayesian Generalized Kernel (BGK) inference has been applied to terrain modeling and traversability assessment. In the work of \cite{shan2018bayesian}, an online traversability analysis method is implemented through two sequential BGK inference steps. While this approach demonstrates improvements in accuracy and efficiency compared to VHR and GPR, it is limited by its reliance on single-frame point clouds, making it susceptible to noise in the data. Additionally, the kernel function used in this method tends to be overly smooth, which hampers its ability to accurately model the terrain discontinuities commonly found in complex environments.

In addition to Bayesian non-parametric inference, terrain modeling can also be approached using Markov Random Fields (MRF) or Conditional Random Fields (CRF). For instance, \cite{zhang2015ground} combine terrain smoothness assumptions with height measurements in a multi-label MRF framework, utilizing loopy belief propagation to achieve robust terrain modeling and ground segmentation.

In another approach, \cite{rummelhard2017ground} model the terrain as a spatio-temporal CRF, estimating terrain elevation using the EM algorithm. \cite{huang2021fast} leverage the graph cut method to address a coarse-to-fine MRF model for effective ground segmentation. Moreover, \cite{forkel2021probabilistic} present a probabilistic terrain estimation method framed as a recursive Gaussian state estimation problem, which they further extend into a dynamic resolution model in~\cite{forkel2022dynamic}. B-spline surfaces are also utilized in traversability analysis due to their excellent capability for fitting complex geometrical shapes, as demonstrated by \cite{rodrigues2020b}. Recent studies have also explored the integration of terrain models with vehicle kinematic models for traversability analysis. \cite{cai2023probabilistic} introduce a probabilistic representation of traversability, modeling it as a distribution that is conditioned on the dynamic behaviors of the robot and the characteristics of the terrain. In a similar vein, \cite{fan2021step} address uncertainties in risk-aware planning, proposing a 2.5D traversability evaluation method that factors in localization errors, sensor noise, vehicle dynamics, terrain characteristics, and various sources of traversability risk. 

\cite{xue2023traversability} present a LiDAR-based terrain modeling approach that fuses multi-frame data to produce accurate and stable terrain models. Using normal distributions transform mapping, it fuses consecutive frames, and applies BGK and bilateral filtering to enhance stability while preserving sharp edges. Traversability is determined through geometric connectivity analysis of the terrain regions. Based on \cite{xue2023traversability}, \cite{xue2023contrastive} propose a self-supervised learning framework for terrain traversability in unstructured environments using contrastive label disambiguation. Pseudo labels are automatically generated by projecting driving experiences onto real-time terrain models. A prototype-based contrastive learning method refines these pseudo labels iteratively, eliminating ambiguities and enabling platform-specific traversability learning without human annotations.

\subsubsection{Semantic Segmentation}

\begin{table*}[htbp]
	\caption{Learning-based semantic segmentation methods for unstructured environments.}
	\begin{center}
		\setlength{\tabcolsep}{1.4mm}{
			\begin{tabular}{c|c|c|c}
				\midrule
				\bf Reference &\bf  Architecture &\bf Sensor & \bf Dataset  \\ 
				\midrule
				\cite{offseg}&BiSeNet v2, HRNet v2+OCR &RGB&RELLIS-3D\, RUGD\ \\
				\cite{singh2021offroadtranseg}&ViT, ResNet50, DINO&RGB&RELLIS-3D, RUGD \\
				\cite{liu2021point}&Euclidean Clustering, RANSAC&LiDAR&Gazebo \\
				\cite{zhou2023off}&PointTensor, Cylinder&LiDAR&RELLIS-3D, CARLA \\
				\cite{viswanath2024off}&FCN&LiDAR&RELLIS-3D \\
				\cite{jiang2023ross}&DeepLab v3&RADAR&RELLIS-3D \\
				\cite{lian2023research}&Patchwork++, LIDROR, YOLO v5&RGB, LiDAR&KITTI, Off-road \\
				\cite{feng2024multi}&RangeNet++, ResNet&RGB, LiDAR& RELLIS-3D, CARLA \\
				\cite{kim2024uncertainty}&BKI, EDL&RGB, LiDAR&RELLIS-3D, Off-road \\
				\cite{li2024unstrprompt}&GPT-3.5, CLIP&RGB&IDD, ORFD, AutoMine\\
				\midrule
			\end{tabular}
			\label{tab:ss_results}
		}
	\end{center}
\end{table*}

\begin{figure*}[h]
	\centering
	\includegraphics[width=0.9\textwidth]{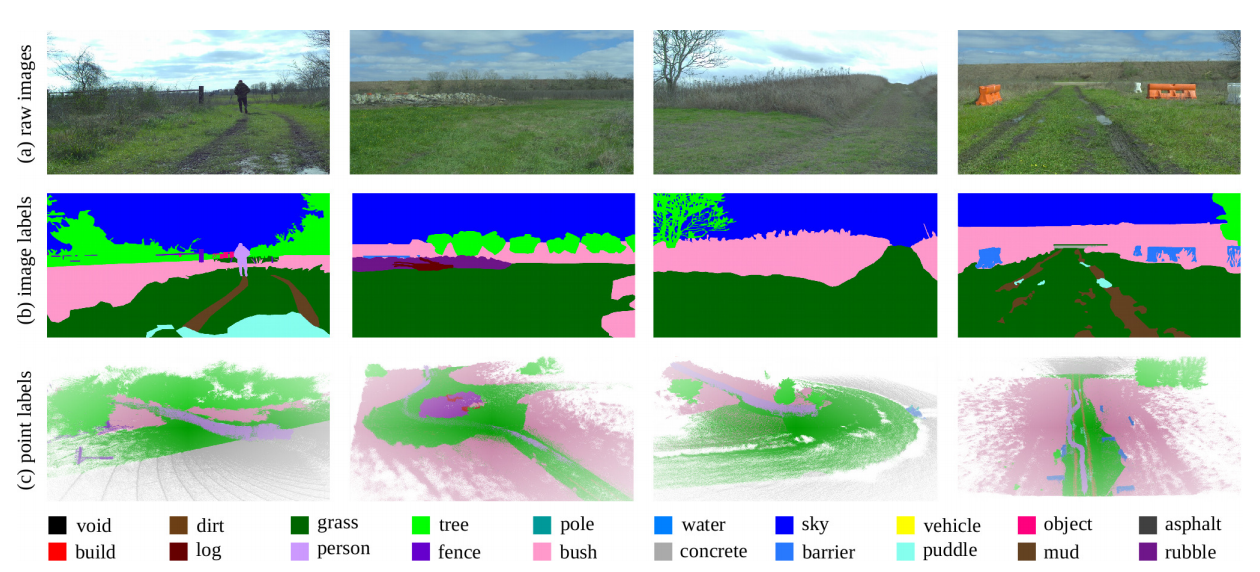}
	\caption{Semantic segmentation in the RELLIS-3D dataset~\cite{rellis3d} collected from unstructured environments.}
	\label{fig:ss}
\end{figure*}

Compared to traversability estimation, semantic segmentation offers a more comprehensive understanding of a vehicle’s surroundings~\cite{yu2018bisenet}, as shown in Fig.~\ref{fig:ss}. With the advancement of deep learning, several semantic segmentation algorithms tailored for unstructured scenarios in autonomous driving have been proposed, drawing inspiration from mature semantic segmentation methods~\cite{ronneberger2015u,chen2014semantic,zhao2017pyramid,cylinder3d}. Detailed information on these methods is provided in Table~\ref{tab:ss_results}.

Some methods use LiDAR point clouds for semantic segmentation. The method in~\cite{liu2021point} combines Euclidean clustering with multi-plane extraction to handle overhanging objects and optimize segmentation time with a multi-resolution grid method.
MAPC-Net~\cite{zhou2023off} is a semantic segmentation network for LiDAR applications, incorporating multi-layer receptive field fusion and gated feature fusion techniques. It utilizes CARLA for dataset generation and augments training data to address sample imbalance.
The study in~\cite{viswanath2024off} improves object segmentation using LiDAR intensity, validated across different LiDAR systems, and integrates calibrated intensity for enhanced prediction accuracy.
RADAR, which can penetrate through dust and haze, is addressed in ROSS~\cite{jiang2023ross}, introducing a pipeline that generates RADAR labels from LiDAR data and demonstrates RADAR's potential in off-road navigation.

\subsection{RGB-based Methods}
\subsubsection{Traversability Estimation}

\begin{table*}[htbp]
	\caption{Vision-based traversability estimation methods for unstructured environments.}
	\begin{center}
		\setlength{\tabcolsep}{1.9mm}{
			\begin{tabular}{c|c|c|c}
				\midrule
				\bf Reference & \bf Architecture &\bf Method &\bf  Dataset  \\ 
				\midrule
				\cite{gao2021fine}&AlexNet &Contrastive Learning&Private data \\
				\cite{gao2022active}&AlexNet &Active and Contrastive Learning&Private data, DeepScene \\
				\cite{sun2023passable}&PSPNet&Supervised Learning&ORFD \\
				\cite{seo2023learning}&PSPNet&Self-Supervised Learning&Private data \\
				\cite{jung2023v}&ViT-H SAM&Self-Supervised Learning& RELLIS-3D \\
				\cite{chung2024pixel}&EfficientNet-B0&Supervised Learning&Private data \\
				\midrule
			\end{tabular}
			\label{tab:img-te}
		}
	\end{center}
\end{table*}
Although LiDAR can accurately capture spatial structural information, its high cost has led to the proposal of many vision-based traversability estimation algorithms~\cite{xiao2016monocular,wang2016multi}. Table~\ref{tab:img-te} lists the specific details of these methods.

\cite{sun2023passable} enhance passable area extraction using a modified PSPNet with MobileNetv2 and dilation convolution modules, improving feature extraction and fusion. The method shows significant improvements in detecting passable areas in off-road environments.
\cite{chung2024pixel} propose a learning-based method for real-time long-range terrain elevation map prediction from onboard egocentric images. It includes a transformer-based encoder, orientation-aware positional encoding, and a history-augmented map embedding to enhance elevation map consistency and performance.

Some algorithms employ contrastive learning by comparing traversable and non-traversable regions.
\cite{gao2021fine} introduce a contrastive learning method for terrain analysis using human-annotated anchor patches to distinguish traversability levels, and develops fine-grained semantic segmentation and mapping techniques. Experimental validation shows effectiveness across varied off-road scenarios.
\cite{gao2022active} propose an active contrastive learning approach using patch-based weak annotations instead of pixelwise labels. It employs contrastive learning for feature representation and adaptive clustering for category discovery, with a risk evaluation technique for prioritizing high-risk frames for labeling.

Some algorithms also adopt a self-supervised approach to reduce the reliance on data labeling.
\cite{seo2023learning} present a self-supervised approach for learning traversability from images, deriving self-supervised labels from historical driving data and training a neural network with a one-class classification algorithm, integrating advanced self-supervised techniques.
\cite{jung2023v} introduce a self-supervised learning method for image-based traversability prediction using contrastive representation learning and instance-based segmentation masks, outperforming recent methods and showing compatibility with a model-predictive controller.

\subsubsection{Semantic Segmentation}
Some methods focus on semantic segmentation for autonomous driving in unstructured scenes using cameras, such as OFFSEG~\cite{offseg} and OffRoadTranSeg~\cite{singh2021offroadtranseg}. OFFSEG~\cite{offseg} introduces a framework for off-road semantic segmentation with two components: (i) a class semantic segmentation that categorizes scenes into four classes and (ii) a color segmentation approach for identifying subclasses within traversable regions. The framework is evaluated on RELLIS-3D~\cite{rellis3d} and RUGD~\cite{rugd} datasets.
OffRoadTranSeg~\cite{singh2021offroadtranseg} uses a semi-supervised approach with a self-supervised vision transformer for off-road segmentation. It leverages self-supervised data collection and demonstrates improved performance on RELLIS-3D and RUGD datasets, addressing class imbalance issues.

\subsection{RADAR-based Methods}
In unstructured off-road environments, perception systems often need to cope with harsh conditions such as dust, fog, rain, snow, and vegetation occlusion, which can severely degrade the performance of conventional vision and LiDAR sensors. RADAR, due to its strong penetration capability and environmental robustness, has gradually become an important perception modality for off-road autonomous driving. Unlike sensors that rely on visible light or laser reflection, RADAR can detect obstacles and estimate relative velocity in low-visibility or occluded scenarios, making it particularly suitable for deserts, farmland, or post-disaster environments. In recent years, millimeter-wave and frequency-modulated continuous-wave (FMCW) RADAR have attracted considerable attention for 3D environment perception and terrain reconstruction. For example, This paper demonstrates that RADAR, adapted from existing LiDAR-based methods, provides longer-range perception, robust ground plane estimation and obstacle detection, and enables successful autonomous navigation in challenging off-road environments.

\subsection{Thermal-based Methods}
Thermal imaging plays an important role in off-road autonomous driving. Compared with traditional RGB vision, thermal sensors can perceive the environment reliably under challenging conditions such as nighttime, glare, dust, fog, or vegetation occlusion. They are independent of visible light, highly resistant to interference, and capable of detecting living beings, obstacles, and terrain boundaries, thereby enhancing safety and robustness in extreme or low-visibility environments. However, thermal images contain limited texture and lower resolution, making them insufficient for fine-grained perception alone. Therefore, they are often fused with RGB or LiDAR data to leverage complementary information across modalities, significantly improving environmental understanding and navigation reliability in off-road scenarios.

\subsection{Fusion-based Methods}


\subsubsection{Traversability Estimation}

Some methods combine LiDAR’s geometric structure with camera texture information, using multi-sensor fusion for traversability estimation~\cite{xiao2018hybrid,xiao2015crf}. Table~\ref{tab:f-te-results} presents the experimental results of fusion-based traversability estimation on the ORFD~\cite{orfd} dataset.

\begin{table*}[h]
	\caption{Quantitative results of fusion-based traversability estimation methods on the testing set of ORFD~\cite{orfd}.}
	\begin{center}
		\setlength{\tabcolsep}{1.0mm}{
			\begin{tabular}{c|ccc|ccccc|cc}
				\midrule
				\multirow{2}{*}{\bf Method} & \multicolumn{3}{c|}{\textbf{Modality}} &\multirow{2}{*}{\bf Accuracy} 
				&\multirow{2}{*}{\bf Precision} &\multirow{2}{*}{\bf  Recall} &\multirow{2}{*}{ \bf F-score} &\multirow{2}{*}{\bf  IOU}&\multirow{2}{*}{\bf \textcolor{red}{Params}}&\multirow{2}{*}{\bf \textcolor{red}{Speed}} \\ 
				&RGB&Depth&Normal&&&&&&& \\
				\midrule
				FuseNet&$\checkmark$&$\checkmark$&&87.4\% & 74.5\% &85.2\%&79.5\%&66.0\%&50.0M&{\bf49.0} Hz\\  
				SNE-RoadSeg&$\checkmark$&&$\checkmark$& 93.8\% &  86.7\% &92.7\%&89.6\%&81.2\%&201.3M&12.5 Hz\\ 
				OFF-Net&$\checkmark$&&$\checkmark$&  94.5\%& 86.6\% &94.3\%&90.3\%&82.3\%&{\bf25.2}M&33.9 Hz\\  
				
				FSN-Swin&$\checkmark$&&$\checkmark$& 96.4\% & 91.2\% &95.8\%&93.4\%&87.7\%&-&-\\ 
				M2F2-Net&$\checkmark$&&$\checkmark$& 98.1\% & 97.3\% &95.5\%&96.4\%&93.1\%&26.8M&34.0 Hz\\
				OFF-CSUNet&$\checkmark$&&$\checkmark$& {\bf98.6}\% &{\bf97.7}\% &95.6\% &96.9\% &{\bf94.1}\% &31.1M &26.0Hz\\ 
				NAIFNet&$\checkmark$&&$\checkmark$&98.4\% & 97.5\% &{\bf96.4}\%&{\bf97.0}\%&{\bf94.1}\%&26.8M&11.5 Hz\\ 
				\midrule
			\end{tabular}
			\label{tab:f-te-results}
		}
	\end{center}
\end{table*}

\cite{guan2021tns} introduce the terrain traversability mapping and navigation system for autonomous excavators in unstructured environments, integrating RGB images and 3D point clouds into a global map for real-time navigation.
VINet~\cite{guan2023vinet} is a visual and inertial-based network for terrain classification in robotic navigation, using a novel labeling scheme for generalization on unfamiliar surfaces and an adaptive control framework for improved navigation performance.
\cite{bae2023self} present a deep metric learning approach that integrates unlabeled data with few prototypes for binary segmentation and traversability regression, introducing a new evaluation metric for comprehensive performance assessment.
\cite{feng2023adaptive} address untrustworthy features in depth images with the adaptive-mask fusion network, which uses adaptive-weight masks to merge RGB and depth images, improving segmentation accuracy.

VPrs-Net~\cite{liu2023novel} is a multi-task network for vanishing point detection and road segmentation, enhancing accuracy in off-road environments with VP-guided models and a multi-attention architecture. Evaluation on the ORFD dataset demonstrates its effectiveness.
M2F2-Net~\cite{ye2023m2f2} and OFF-CSUNet~\cite{OFF-CSUNet} are multi-modal networks for free-space detection, using multi-modal cross fusion and a multi-modal segmentation decoder to integrate RGB and surface normal maps, including an edge segmentation decoder module for improved road edge extraction.
\cite{duraisamy2023multi} introduce a framework for classifying and clustering regions, incorporating LiDAR-based ground segmentation with adaptive likelihood estimation and late fusion of deep learning and LiDAR results for improved drivable region classification.
\cite{xu2023unstructured} combine terrain features from 3D point clouds and RGB images to create a global terrain traversal cost map, using Bayesian generalized kernel inference and Kalman filtering for real-time local elevation maps and probabilistic semantic mapping.

UFO~\cite{kim2024ufo} is a learning-based fusion approach for dense terrain classification maps using LiDAR and image data, incorporating uncertainty-aware pseudo-labels to improve robustness and accuracy in off-road environments.
RoadRunner~\cite{frey2024roadrunner} is a self-supervised framework for predicting terrain traversability and generating elevation maps from camera and LiDAR inputs, using an end-to-end approach with sensor fusion for enhanced accuracy in terrain assessment.
NAIFNet~\cite{lv2024noise} is designed for off-road free-space detection, incorporating a noise-aware intermediary interaction module and a denoising-guided decoder to address noise interference during multimodal fusion.
FNS-Swin~\cite{yan2024fsn} enhances Transformer networks for segmentation tasks by integrating data fusion, using the Swin Transformer's sliding window structure and cross-attention mechanisms to improve feature correlation and depth information extraction.

\subsubsection{Semantic Segmentation}
Integrating image and LiDAR data for better scene perception is explored in~\cite{lian2023research}, which includes ground point segmentation, noise filtering, and improved Euclidean clustering for obstacle detection.
The MF-SN Net~\cite{feng2024multi} proposes a multisensor fusion network for unstructured scene segmentation, incorporating surface normal information and novel methods for feature reweighting and cross-layer attention.
In~\cite{kim2024uncertainty}, an evidential semantic mapping framework is introduced, utilizing dempster-shafer theory of evidence for uncertainty management in mapping.
Recent advancements include UnstrPrompt~\cite{li2024unstrprompt}, a language prompt set for unstructured environments derived from prominent datasets and featuring a structured prompt generation approach.
\section{Path Planning}
\label{planning}

\begin{figure*}[h]
	\centering
	\includegraphics[width=0.9\textwidth]{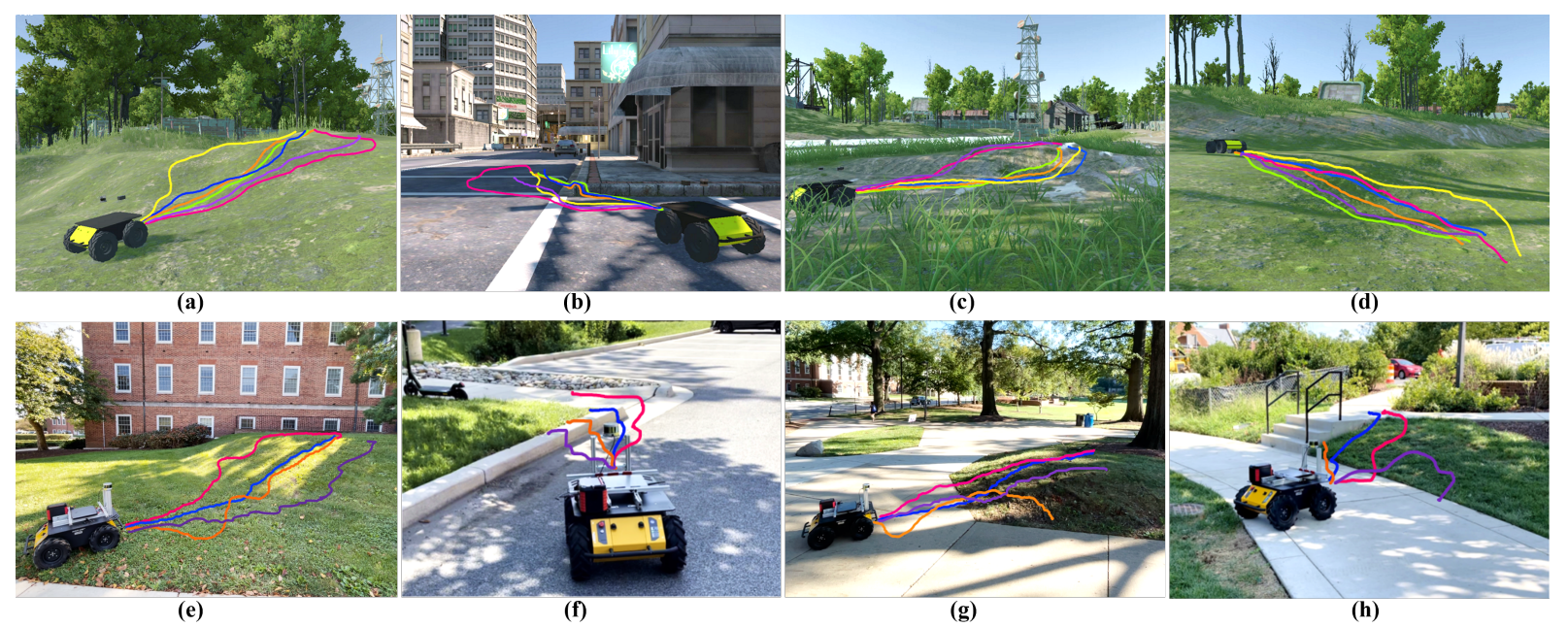}
	\caption{Local path planning when navigating in different unstructured environments~\cite{weerakoon2022terp}.}
	\label{fig:pp}
\end{figure*}


Path planning~\cite{gasparetto2015path} and trajectory planning~\cite{shang2020guide} algorithms are processes that create the optimal route to successfully reach the destination through navigable areas~\cite{wang2021multiple}. This optimality can be reflected in path length, energy consumption, or any other metrics deemed important by the operator~\cite{barthlow2020off}. Path planning methods are divided into global path planning and local path planning (Fig.~\ref{fig:pp}). Because of the diversity and complexity of actual application scenarios, the requirements for path planning vary in different scenarios, so there are many types of path planning methods, as Table~\ref{tab:pp} shows. 

\begin{table*}[h]
	\centering
	\caption{The advantage and disadvantage of different path planning methods.}
	\begin{center}
		\begin{tabular}{p{1.2cm}|p{3.6cm}|p{5.5cm}|p{4.5cm}}
			\midrule
			 &\bf Type& \bf Advantage & \bf Disadvantage\\
			\midrule
			\multirow{12}{*}{Global}&\multirow{4}{*}{Search-based} & Excellent performance in handling static environments, small-scale maps, and problems with high requirements for optimality & Computational efficiency is limited when dealing with high-dimensional spaces and large-scale environments  \\
			\cline{2-4}
			&\multirow{5}{*}{Sampling-based} & Suitable for high-dimensional spaces and dynamic environments, can quickly generate feasible paths, and has advantages in complex and uncertain environments & Cannot guarantee that the generated path is optimal   \\
			\cline{2-4}
			&\multirow{4}{*}{Biologically-inspired} & Good adaptability for complex multi-objective optimization and large-scale search spaces, and can discover novel paths & Computational time may be longer   \\
			\midrule
			\multirow{14}{*}{Local}&\multirow{4}{*}{Optimization-based} & Can precisely model for specific optimization goals and constraints, and provide accurate solutions when considering specific goals & May have higher computational complexity  \\
			\cline{2-4}
			&\multirow{3}{*}{Artificial Potential Field} & Simple calculation, good real-time performance, and suitable for local path planning & May fall into local minima and fail to reach the target point   \\
			\cline{2-4}
			&\multirow{3}{*}{Dynamic Window} & Quick response in dynamic environments and can handle emergencies & Sensitive to parameter settings, and the path may not be smooth   \\
			\cline{2-4}		
			&\multirow{4}{*}{Data-driven} & Outstanding performance in handling environments with large amounts of data and complex patterns, and can automatically learn & Requires a large amount of data for training, and the model interpretability is weak   \\
			\midrule
		\end{tabular}
		\label{tab:pp}
	\end{center}
\end{table*}

Compared to structured environments, path planning in unstructured environments faces several major challenges. These challenges include high uncertainty and dynamic changes, complex state spaces, limited prior knowledge, real-time operation requirements, and the complexity of long-term planning~\cite{chu2012local,erke2020improved,guo2023survey}. This section will discuss these challenges in detail and explore path planning algorithms for unstructured environments.

\subsection{Challenges of Path Planning in Unstructured Environment}

\subsubsection{Challenges of Global Path Planning}
In autonomous driving systems, the global path planning module can generate a path to guide the autonomous vehicle based on prior information about the target environment and the specific task requirements of the vehicle. However, when navigating complex and unknown unstructured environments, the lack of prior information severely limits the functionality of the global path planning module.
\begin{itemize} 
	
	\item \textbf{Satellite Imagery as the Sole Data Source}: In unknown unstructured environments, the lack of high-precision prior maps means that prior information about the target scene can only be obtained through satellite imagery. As a result, the global path planning module relies solely on satellite data to generate the global guidance path. However, the paths generated in this manner often have significant inaccuracies, with precision levels that fall far short of meeting the requirements for subsequent local path planning modules.
\end{itemize}
\subsubsection{Challenges of Local Path Planning}
In autonomous driving systems, the local path planning module integrates global guidance information, pose estimation data, and environmental perception to dynamically generate the local driving path for the vehicle over a short time horizon. However, when faced with complex and unknown unstructured environments, the local path planning module encounters several challenges.
\begin{itemize} 
	
	\item \textbf{Highly Sensitive to Positioning Errors}: In unstructured environments, accurate and robust positioning of the vehicle is particularly challenging due to the limited availability of reliable positioning methods. The local path planning module is often highly sensitive to positioning errors. If the pose estimation module fails or produces inaccurate results, the planned local path can deviate significantly, increasing the risk of incorrect routes and improper driving behavior, which poses a serious threat to the safety of autonomous driving.
	\item \textbf{Dependence on Geometric Constraints}: Existing local path planning algorithms tend to heavily rely on geometric constraint information. However, in unstructured environments, natural terrain features dominate, providing no clear or orderly constraint information for the unmanned platform. Moreover, these natural elements often lack spatial regularity, rendering geometric constraints either disordered or ineffective, which significantly increases the complexity and uncertainty of the local path planning process. For example, in sparsely featured environments such as deserts or grasslands, the scarcity of geometric constraint information leaves the local path planning module with insufficient decision-making support. Conversely, in densely featured environments like jungles or mountainous areas, the abundance of irregularly distributed features results in overly redundant geometric constraints, making it difficult for the local path planning module to generate feasible local paths.
\end{itemize}

%
%
%
%
%
%
%
%

Common global path planning algorithms include the A* algorithm~\cite{candra2020dijkstra}, Dijkstra's algorithm~\cite{wang2011application}, and Rapidly-exploring Random Trees (RRT)~\cite{lavalle2001rapidly}. Local path planning focuses on immediate path adjustments during the robot's movement, considering the environmental information around the robot's current position. Common local path planning algorithms include the Dynamic Window Approach (DWA)~\cite{fox1997dynamic}, the Optimization-based method~\cite{4669725}, and the Artificial Potential Field (APF)~\cite{warren1989global} method. Data-driven methods can be used for both global and local path planning.

\subsection{Global Path Planning}

\subsubsection{Search-based Methods}

The main idea behind search-based path planning algorithms~\cite{sanchez2021path} is to preprocess the planning space into a grid map containing obstacle information. In the grid map, nodes represent discrete states in the state space, and state transitions are achieved through connections between nodes. Compared to other path planning algorithms, search-based algorithms have higher solution stability, consistently producing the same planning results in identical scenarios. However, due to the discretization of the continuous planning space, the generated paths may not be smooth enough, introducing quantization errors. Table~\ref{tab:Search} presents the search-based path planning algorithms for unstructured environments.

\begin{table}[htbp]
	\caption{Search-based path planning methods for unstructured environments.}
	\begin{center}
		\resizebox{0.6\textwidth}{!}
		{
			\begin{tabular}{c | c }
				\midrule
				\bf Reference & \bf Algorithm \\ \midrule
				\cite{stentz1994optimal} &Dynamic A* \\ 
				\cite{likhachev2005anytime} &Anytime Dynamic A*\\ 
				\cite{montemerlo2008junior} &AD*\\ 
				\cite{ziegler2008navigating} & A*, Voronoi Cost Function\\ 
				\cite{li2012ara} &Anytime Repairing A* \\ 
				\cite{roy2018hierarchical} &Dynamic Programming \\ 
				\cite{sedighi2019guided} & Hybrid A*\\ 
				\cite{sakayori2021energy} &Dijkstra's Algorithm \\ 
				\cite{hong2021improved} &Improved A*\\ 
				
				\cite{hua2022global} &PRM, Improved A*\\ 
				\cite{xu2024multi} &Velocity Hybrid A*\\ 
				\midrule
			\end{tabular}
			\label{tab:Search}
		}
	\end{center}
\end{table}

Dijkstra's algorithm, an early search algorithm, is mainly used to find the shortest path between any two nodes in a connected graph. It adopts a breadth-first search approach, starting from the initial node and progressively evaluating nodes adjacent to the current node. Its search strategy is greedy, defining a cost function (usually the distance to the starting point) and selecting the node with the lowest cost each time to include in the path. If a feasible path exists between the start and end points, this algorithm guarantees finding the shortest feasible path. \cite{sakayori2021energy} considere the power consumption model of planetary rovers on rugged terrain, designed a cost function accounting for both generation and consumption, and used Dijkstra's algorithm to obtain path planning results. However, as a breadth-first search method, when the search range needs to be expanded or finer gridding is required, the number of nodes in the map will increase significantly, leading to a substantial increase in the total search volume of the algorithm and a rapid growth in the time to solve the optimal path. Therefore, Dijkstra's algorithm is rarely directly used for large-scale or high-precision path planning problems.

Dynamic Programming (DP) solves optimal solutions by breaking problems into smaller subproblems. Its main idea is to utilize the overlapping nature of subproblems, reducing computational redundancy through memorization or recursion, significantly improving computational efficiency. \cite{roy2018hierarchical} propose an efficient hierarchical method for vehicle path planning in vast off-road terrains. Global path planning uses the dynamic programming algorithm based on available low-resolution terrain information before the mission to find the optimal path outline. Subsequently, the local optimization program fine-tunes the optimal predefined route based on the results of the global program and integrates high-resolution terrain maps and obstacle detection algorithms.

The A* algorithm is a classic heuristic search algorithm, particularly suitable for discrete grid maps. It combines the advantages of best-first search and Dijkstra's algorithm, finding the path with the lowest cost by evaluating the path cost and heuristic value (usually the straight-line distance from the current node to the target node). The A* algorithm's path search process includes the following steps:

\textbf{Initialization}: Add the start node to the open list, and initialize the closed list as empty.

\textbf{Node Selection}: Select the node with the lowest \( f(n) = g(n) + h(n) \) value from the open list, where \( g(n) \) is the actual cost from the start node to node \( n \), and \( h(n) \) is the estimated cost from node \( n \) to the target node (heuristic function).

\textbf{Node Expansion}: Move the current node from the open list to the closed list. For each neighboring node of the current node, if the neighbor node is not in the closed list, calculate its \( f(n) \) value. If the neighbor node is not in the open list or the new \( f(n) \) value is lower, update its parent node to the current node and add or update it in the open list.

\textbf{Termination Condition}: The algorithm terminates when the target node is added to the closed list or the open list is empty. If the target node is in the closed list, the path has been found; otherwise, the path does not exist.

Improved versions of the A* algorithm include Dynamic A*~\cite{stentz1994optimal}, Anytime Repairing A*\cite{li2012ara}, Anytime Dynamic A*~\cite{likhachev2005anytime}, and Hybrid A*~\cite{sedighi2019guided}. In the 2007 DARPA Urban Challenge, the vehicle Boss, using the AD* algorithm, achieved great success, demonstrating its practical application~\cite{montemerlo2008junior}. The A* algorithm improves search efficiency through heuristic functions, performing exceptionally well in large-scale graph searches. \cite{hong2021improved} propose an improved A* algorithm based on terrain data for long-distance path planning tasks. This method uses terrain data maps generated from digital elevation models and optimizes the algorithm in data structure and retrieval strategy. As long as the heuristic function is reasonable (i.e., never overestimates the actual cost), the A* algorithm guarantees finding the optimal path. The heuristic function can be adjusted according to specific applications to meet different path planning needs.

\cite{ziegler2008navigating} implement the A* algorithm combitned with a Voronoi cost function for planning in unstructured spaces and parking lots. However, in complex or large-scale search spaces, the A* algorithm may face high time and space complexity, consuming significant computational resources. Additionally, selecting an appropriate heuristic function is crucial for the algorithm's performance, as an unsuitable choice may lead to inefficient searches. \cite{stentz1994optimal} proposes a new algorithm, D*, capable of planning paths efficiently, optimally, and completely in unknown, partially known, and changing environments. 

\cite{hua2022global} propose a path planning method considering terrain factors and soil mechanics' impact on UGV maneuverability. They used the probabilistic roadmap method to establish a connection matrix and a multidimensional traffic cost evaluation matrix between sampling nodes and proposed an improved A* algorithm based on vehicle mobility costs to generate global paths. \cite{xu2024multi} propose a centralized decision-making distributed planning framework to solve the Multi-Vehicle Trajectory Planning (MVTP) problem in unstructured conflict areas, using Velocity Hybrid A* (V-Hybrid A*) for trajectory search.

\subsubsection{Sample-based Methods}

Sampling-based path planning algorithms typically have probabilistic completeness, meaning they can find the optimal path that exists on the planning map given a sufficient number of samples. These methods introduce randomness into the path planning process, with the most common including the Probabilistic Roadmap (PRM)~\cite{kavraki1998analysis} and the RRT methods. Table~\ref{tab:sample} lists the sample-based path planning algorithms for unstructured environments.

\begin{table}[htbp]
	\caption{Sample-based path planning methods for unstructured environments.}
	\begin{center}
		\begin{tabular}{c | c }
			\midrule
			\bf Reference & \bf Algorithm \\ \midrule
			\cite{jaillet2010sampling} &T-RRT \\ 
			\cite{jiang2021r2} &R2-RRT* \\ 
			\cite{yin2023efficient} &ER-RRT*, Surrogate Modeling\\
			\cite{tian2023driving} &Potential Field-based RRT* \\
			\cite{zheng2024two} &PRM,  A*\\ 
			\midrule
		\end{tabular}
		\label{tab:sample}
	\end{center}
\end{table}

The PRM algorithm first randomly selects nodes in the search space, then connects all nodes, including the start and end points, and performs collision detection to form a collision-free connected graph. Finally, it uses a search algorithm to find the shortest path connecting the start and end points. The PRM algorithm also has probabilistic completeness and avoids a complete search of the entire area through sampling, thereby reducing the computational load. Thus, this algorithm is more efficient and simpler to implement than grid search methods, and it is widely used in high-dimensional space planning problems. \cite{zheng2024two} propose a two-stage path planning algorithm for complex off-road environments. They used an improved PRM algorithm to quickly and roughly determine the initial path, and the A* algorithm for precise path planning.

The RRT algorithm is a sampling-based path planning method, especially suitable for path search problems in high-dimensional spaces. Its core idea is to quickly expand a tree structure through random sampling to find a feasible path from the starting point to the target point. The RRT algorithm is simple to operate, has fast solution speed, and good real-time performance, and it is widely used in high-dimensional spaces or large-scale maps. \cite{jaillet2010sampling} propose T-RRT, which can generate not only feasible but also high-quality paths. T-RRT considers a user-given cost function defined over the configuration space as an additional input to the standard path-planning problem. For example, in outdoor navigation problems, the cost map may correspond to the elevation map of the terrain to compute movements that minimize climbs in steep slope areas.

The RRT*~\cite{xu2024recent} algorithm is an improved version of the RRT algorithm, aimed at addressing the issue of low-quality paths generated by RRT. RRT* introduces a path optimization mechanism, ensuring path feasibility while seeking paths closer to optimal. Its basic steps are as follows:

\textbf{(1) Initialization}: Set the starting point as the root node of the tree and initialize the tree \( T \).

\textbf{(2) Random Sampling}: Randomly generate a sample point \( x_{rand} \) in the search space.

\textbf{(3) Tree Expansion}: Find the node \( x_{nearest} \) in the tree that is closest to \( x_{rand} \). Move a certain distance toward \( x_{rand} \) to generate a new node \( x_{new} \), and check if the path from \( x_{nearest} \) to \( x_{new} \) is collision-free. If the path is feasible, add \( x_{new} \) to the tree.

\textbf{(4) Reconnection Optimization}: In the neighborhood of \( x_{new} \), find all nodes \( x_{near} \) and calculate the path cost from these nodes through \( x_{new} \) to the goal. If the path cost through \( x_{new} \) to the goal is lower than the direct path cost from these nodes to the goal, update the parent node of these nodes to \( x_{new} \) to optimize the path.

\textbf{(5) Iterative Update}: Repeat steps 2-4 until \( x_{new} \) reaches the target area or the preset number of iterations is reached.

Through path reconnection and optimization mechanisms, the RRT* algorithm can find near-optimal paths in the search space. \cite{tian2023driving} propose a potential field-based RRT* motion planning algorithm for vehicles to avoid risks, constructing a configuration space with potential fields to determine risk ranges around obstacles and off-road terrain. Due to its random sampling characteristics, the RRT* algorithm has global search capabilities, avoiding local optima. However, the optimization process requires additional computation time, and the complexity of nearest node and path feasibility detection may result in less ideal performance in applications with high real-time requirements. \cite{jiang2021r2} propose state mobility reliability and task mobility reliability to quantify the mobility reliability of AGVs. They developed two reliability-based robust mission planning models and generated optimal paths that meet specific reliability requirements. \cite{yin2023efficient} combine surrogate modeling with the RRT* algorithm and proposed a novel and efficient reliability-based global path planning method.

\subsubsection{Biologically-inspired Methods}

Bio-inspired path planning methods mimic the behavior and decision-making processes of biological organisms in natural environments to tackle path planning challenges in complex environments. Common bio-inspired algorithms include Ant Colony Optimization (ACO)~\cite{dorigo2006ant}, Particle Swarm Optimization (PSO)~\cite{kennedy1995particle}, and Genetic Algorithms (GA)~\cite{mitchell1998introduction}. These algorithms utilize strategies observed in natural selection and evolution to find optimal paths.

\begin{table}[htbp]
	\caption{Biologically-inspired path planning methods for unstructured environments.}
	\begin{center}
		\begin{tabular}{c | c }
			\midrule
			\bf Reference & \bf Algorithm \\ 
			\midrule
			\cite{wang2018off} &ACO \\ 
			\cite{zhu2020path} &ACO, APF \\ 
			\cite{peng2021multiobjective} &GA, PSO \\ 
			\cite{hu2022novel} &Knowledge-based GA\\ 
			\cite{li2023research} &Improved GA \\ 
			\midrule
		\end{tabular}
		\label{tab:Bio}
	\end{center}
\end{table}

\textbf{Ant Colony Optimization}: Simulates the foraging behavior of ants, where ants leave pheromone trails to mark paths and use positive feedback to find the optimal path. As ants move, they deposit pheromones, which evaporate over time; paths with higher pheromone concentrations attract more ants, gradually forming the optimal path. Table~\ref{tab:Bio} details the biologically-inspired path planning methods for unstructured environments. \cite{wang2018off} propose an improved ant colony path planning algorithm after analyzing the combined impact of terrain slope and soil strength on vehicle off-road passability. \cite{zhu2020path} combine the artificial potential field method with ACO, introducing inducible heuristic factors and dynamically adjusting the state transition rules of ACO, resulting in higher global search capabilities and faster convergence.

\textbf{Particle Swarm Optimization}: Simulates the behavior of bird flocks, finding the optimal solution through cooperation and competition among particles. Each particle represents a candidate solution, adjusting its position based on its own experience and the experience of the swarm, gradually converging to the optimal solution. \cite{peng2021multiobjective} represent vehicle ride comfort, road-holding, and handling performance in rugged terrain through the weighted root mean square value of sprung mass vertical acceleration, suspension travel, and tire deformation, optimizing these objectives using GA, PSO, and a GA-PSO hybrid algorithm.

\textbf{Genetic Algorithm}: Simulates natural selection and genetic variation processes, evolving to the optimal solution through iterative selection, recombination, and mutation of individuals in the population. Individuals in the population represent candidate solutions, with higher fitness individuals more likely to reproduce and pass on their characteristics to the next generation. \cite{hu2022novel} propose a novel knowledge-based genetic algorithm for robot path planning in complex unstructured environments, developing multiple operators to enhance the algorithm's performance, such as combining local path planning operators. \cite{li2023research} propose an improved genetic algorithm to address the shortcomings of basic genetic algorithms.

Bio-inspired algorithms have strong global search capabilities and effectively avoid local optima. They are highly adaptable to environmental changes and uncertainties, capable of handling complex and dynamic unstructured environments. The process of bio-inspired algorithms begins with environment modeling, discretizing terrain and obstacle data to create a manageable search space. Then, an appropriate bio-inspired method is selected, such as ACO, PSO, or GA, defining algorithm parameters according to specific application needs, such as pheromone evaporation rate (ACO), inertia weight (PSO), and crossover/mutation rates (GA). Next, the selected algorithm is executed to search for the optimal path, avoiding obstacles while satisfying constraints such as path length and smoothness. Finally, the found path is post-processed to improve its smoothness and feasibility. However, due to the need to simulate the behavior of many individuals or particles, these algorithms may have high computational demands and face challenges in real-time efficiency.

\subsection{Local Path Planning}
\subsubsection{Optimization-based methods}

Optimization-based methods transform path planning problems into mathematical optimization problems by defining an objective function (such as path length, energy consumption, time, etc.) and constraints (such as avoiding obstacles, path smoothness, etc.), and then use optimization algorithms to find the optimal path. In many cases, optimization methods can find the global optimum or a solution close to it. The objective function and constraints can be adjusted according to specific problem requirements, providing great flexibility. However, the computational complexity of solving complex optimization problems can be high, especially for nonlinear and mixed integer optimization problems. Common optimization algorithms include Linear Programming (LP)\cite{dantzig2002linear}, Nonlinear Programming (NLP)\cite{bertsekas1997nonlinear}, and Mixed Integer Programming (MIP)\cite{achterberg2013mixed}, as shown in Table~\ref{tab:opti}.

\begin{table}[htbp]
	\caption{Optimization-based methods for unstructured environments.}
	\begin{center}
		\begin{tabular}{c | c }
			\midrule
			\bf Reference & \bf Algorithm \\ \midrule
			\cite{4669725} &SNOPT,FilMINT, MINLP BB\\ 
			\cite{9744540} &NLP, CPLEX,  A* \\ 
			\cite{10488684} &MIQP, BFS, STMC \\
			\midrule
		\end{tabular}
		\label{tab:opti}
	\end{center}
\end{table}

Linear Programming: Suitable for optimization problems where both the objective function and constraints are linear, commonly solved using the Simplex method or Interior Point methods.

Nonlinear Programming: Solves optimization problems that contain nonlinear terms in the objective function or constraints, typically addressed with Gradient Descent, Newton's method, and other techniques.

Mixed Integer Programming: Solves optimization problems with some variables constrained to be integers, usually solved through Branch and Bound, Cutting Plane methods, and others. MINLP (Mixed-Integer Nonlinear Programming)~\cite{kronqvist2019review}, based on MIP, allows the objective function and constraint conditions to contain nonlinear terms, making the problem more complex. MIQP (Mixed-Integer Quadratic Programming)~\cite{wu2018miqp} is a special case of MIP, in which the objective function is a quadratic function, and the constraint conditions can be linear or quadratic.

In unstructured environment path planning, optimization-based methods can effectively handle complex terrain and obstacle information. During the path planning process, the terrain and obstacle information in the unstructured environment are discretized to form an optimization model that can be processed. Then, based on actual needs, the objective function for path planning is defined, such as the shortest path or minimum energy consumption, and constraints are set, such as avoiding obstacles, maintaining path smoothness, and complying with maximum turning angle limitations. An appropriate optimization algorithm is selected, and the optimal path is found by solving the objective function and constraints. Finally, post-processing is performed on the found path to enhance its smoothness and practical feasibility.

\cite{4669725} develop heuristics for finding good starting points in solving large-scale nonlinear constrained optimization problems, especially NLP and MINLP with nonlinear non-convex functions. The experimental evaluations on NLP and MINLP benchmark problems show that their approach can solve more problems with fewer iterations compared to the best existing solvers from their default starting points.

\cite{9744540} propose a down-sized initialization strategy for optimization-based unstructured trajectory planning, which only optimizes critical variables to provide a proper initial guess for the trajectory planning problem, and they demonstrate that this strategy can facilitate the solution process of the planner and improve the solution time and trajectory quality through experiments.

\cite{10488684} propose a coordinated behavior planning and trajectory planning framework for multiple unmanned ground vehicles in unstructured narrow interaction scenarios, which integrates an efficient behavior planning layer based on MIQP and BFS and an optimal trajectory planning layer with a spatio-temporal motion corridor approach, and they validate the framework in simulations and real-vehicle experiments, demonstrating that it can generate safe, smooth, and efficient trajectories compared to existing planning methods.

\subsubsection{Artificial Potential Field}

The artificial potential field method is based on the concept of fields in physics, primarily involving attractive and repulsive fields. The target position generates an attractive field that pulls the robot towards the target, while obstacles generate a repulsive field that forces the robot to avoid them. The specific principles are as follows:

\begin{table}[htbp]
	\caption{Artificial potential field-based path planning for unstructured environments.}
	\begin{center}
		\resizebox{0.7\textwidth}{!}
		{
			\begin{tabular}{c | c }
				\midrule
				\bf Reference &\bf  Algorithm \\ \midrule
				\cite{yao2020path} &Improved Black Hole Potential Fields, RL \\
				\cite{jiang2024risk} &APF, Coarse2fine A* \\ 
				\cite{linghong2024risk} &APF \\ 
				\midrule
				
			\end{tabular}
			\label{tab:apf}
		}
	\end{center}
\end{table}

\textbf{Attractive Field}: The target position generates an attractive force that draws the robot towards the target. The magnitude of the attractive force is proportional to the distance between the robot and the target; the further the distance, the greater the attractive force.
\begin{eqnarray}
	F_{\text{att}}(q) = -k_{\text{att}} (q - q_{\text{goal}}),
\end{eqnarray}
here, \( F_{\text{att}}(q) \) is the attractive force, \( k_{\text{att}} \) is the attractive coefficient, \( q \) is the robot's current position, and \( q_{\text{goal}} \) is the target position.

\textbf{Repulsive Field}: Obstacles generate a repulsive force that pushes the robot away from the obstacles. The magnitude of the repulsive force is inversely proportional to the distance between the robot and the obstacles; the closer the distance, the greater the repulsive force.
\begin{eqnarray}
	F_{\text{rep}}(q) = 
	\begin{cases} 
		k_{\text{rep}} \left( \frac{1}{q} - \frac{1}{q_{\text{obs}}} \right) \frac{1}{q^2} & \text{if } q \leq q_{\text{obs}} \\ 
		0 & \text{if } q > q_{\text{obs}},
	\end{cases} 
\end{eqnarray}
here, \( F_{\text{rep}}(q) \) is the repulsive force, \( k_{\text{rep}} \) is the repulsive coefficient, \( q \) is the robot's current position, and \( q_{\text{obs}} \) is the influence range of the obstacle.

\textbf{Combined Force}: The robot moves under the combined influence of the attractive and repulsive forces.
\begin{eqnarray}
	F(q) = F_{\text{att}}(q) + F_{\text{rep}}(q).
\end{eqnarray}

The artificial potential field method is relatively simple to implement, has low computational overhead, and thus is efficient. Additionally, since the attractive and repulsive fields are manually designed, it offers strong scalability, allowing multiple factors to be considered during planning. Table~\ref{tab:apf} lists the artificial potential field-based path planning methods for unstructured environments. \cite{jiang2024risk} adopt the concept of artificial potential fields to characterize various risks and incorporate the risk assessment results into the planning module of unmanned ground vehicles. Based on this, they proposed a global planning algorithm called Coarse2fine A*, which considers risks and improves efficiency and flexibility. 

\cite{yao2020path} combine improved black hole potential fields with reinforcement learning, where the black hole potential field is used as the environment in the reinforcement learning algorithm. The agent automatically adapts to the environment and learns how to use basic environmental information to find the target. However, the main limitation of this method is its tendency to get trapped in local optima. This means that during the potential energy descent search process, the algorithm may stall at a local minimum and fail to reach the goal. Furthermore, if an obstacle is near the goal, its repulsive field may interfere with the potential energy of the goal, causing the goal not to be a global minimum and leading to incorrect goal identification. \cite{linghong2024risk} construct driving safety risks in complex and variable urban tunnels based on the theory of artificial potential fields.

\subsubsection{Dynamic Window Approach-based Methods}
\begin{table}[htbp]
	\caption{Dynamic window approach-based path planning methods for unstructured environments.}
	\begin{center}
		\begin{tabular}{c | c }
			\midrule
			\bf Reference & \bf Algorithm \\ \midrule
			\cite{yang2022automatic} &DWA, Improved A*\\
			\cite{eirale2023rl} &DWA, DRL \\ 
			\cite{zhang2024hierarchical} &DWA, Improved D* Lite Algorithm \\ 
			\midrule
		\end{tabular}
		\label{tab:dwa}
	\end{center}
\end{table}

The Dynamic Window Approach (DWA) is based on generating a set of potential velocity pairs (including linear and angular velocities) within the robot's motion space and selecting the optimal pair that allows the robot to get as close as possible to the target while ensuring safety.

The specific implementation process is as follows: First, a set of possible linear and angular velocity pairs is sampled based on the robot's current speed. Then, considering the robot's kinematic constraints and current speed, a dynamic window is formed. The velocity pairs within this window ensure that the robot's movement in the next time step is both safe and feasible. Next, an evaluation function is defined to assess the quality of each velocity pair, typically including metrics such as distance to the target, distance to obstacles, and alignment of velocities. Finally, the evaluation function calculates a score for each velocity pair, and the pair with the highest score is selected as the robot's current motion command.

The DWA algorithm can quickly compute the optimal path in dynamically changing environments. The obstacle distance metric in the evaluation function effectively prevents collisions, ensuring the safety of the path planning. Table~\ref{tab:dwa} details the path planning methods based on the dynamic window approach for unstructured environments. \cite{yang2022automatic} propose an improved A* algorithm and DWA to enhance the accuracy and speed of tracked vehicle automatic parking. However, due to its local search nature, the DWA algorithm may fall into local optima and thus cannot guarantee finding the globally optimal path. \cite{eirale2023rl} also combine deep reinforcement learning with DWA, introducing a new person-following method where the robot's linear velocity was calculated by DWA, and the angular velocity was predicted by the deep reinforcement learning algorithm. \cite{zhang2024hierarchical} use an improved D* Lite algorithm to generate global waypoints for path planning and employed DWA for local path planning, where DWA parameters were selected by a deep reinforcement learning algorithm.

\subsubsection{Data-driven Methods}

Data-driven methods in path planning leverage the power of large datasets to enhance decision-making in navigation environments. These methods achieve high precision and adaptability by learning complex patterns and relationships from extensive data. However, their successful implementation requires careful management of data quality, computational resources, and model complexity to ensure robust performance across various real-world applications.

\begin{table}[htbp]
	\caption{Data-driven path planning methods for unstructured environments.}
	\begin{center}
		\resizebox{0.7\textwidth}{!}{
			\begin{tabular}{c | c }
				\midrule
				\bf Reference & \bf Algorithm \\ 
				\midrule
				\cite{kaushik2018learning} &DDPG \\ 
				\cite{liu2020mapper} &  Evolutionary RL    \\ 
				\cite{semnani2020multi} &GA3C-CADRL, FMP\\
				\cite{ahn2022vision} &Imitation Learning \\ 
				\cite{wang2023deep} &PPO, Curriculum Learning\\ 
				\cite{li2023trajectory} &Deep Learning, Quadratic Optimization \\ 
				\cite{ginerica2024vision} &RNN, DWA \\ 
				\midrule
			\end{tabular}
			\label{tab:Data-driven}
		}
	\end{center}
\end{table}

Learning-based path planning in unstructured environments utilizes neural networks' efficient learning and fitting capabilities in complex environments. Deep neural network models extract features from sensor data (e.g., Cameras, LiDAR) and learn effective path planning strategies~\cite{lecun2015deep}. For instance, CNNs~\cite{gu2018recent} can extract terrain features, recognize obstacles, and identify roads, aiding the robot in determining the optimal path. Deep learning models handle complex unstructured environments, such as mountains and forests, by learning path planning directly from sensor data without the need for manual feature engineering. They can be adjusted and optimized according to different environments and tasks. However, deep learning models typically require a large amount of labeled data for training, especially in complex environments. Training and inference of deep learning models demand substantial computational resources, posing challenges in real-time applications. Moreover, their generalization ability in unseen environments or extreme conditions may be limited. Table~\ref{tab:Data-driven} lists the data-driven path planning methods for unstructured environments.

\cite{ahn2022vision} propose a method for autonomous driving in unstructured environments using vision-based occupancy grid maps, achieved through imitation learning with expert driving data to train deep neural networks. \cite{li2023trajectory} demonstrate the potential of a trajectory planning method based on deep learning and optimization in the practical application of autonomous driving in mining areas. \cite{ginerica2024vision} also propose a vision dynamics approach for path planning and navigation of quadruped robots in unstructured environments, specifically forest roads, using a recurrent neural network that employs RGB-D sensors to construct a sequence of previous depth sensor observations and predict future observations over a limited time span. \cite{xu2022trajectory} propose an end-to-end transformer networks based approach for map-less autonomous driving. The proposed model takes raw LiDAR data and noisy topometric map as input and produces precise local trajectory for navigation. It demonstrates the effectiveness in real-world driving data, including both urban and rural areas. 

Deep reinforcement learning (DRL)~\cite{wang2022deep} in path planning combines deep learning and reinforcement learning techniques to enable agents to autonomously learn optimal paths in complex environments. Specifically, DRL uses neural networks to handle high-dimensional state spaces. The input to the neural network includes various environmental information, such as obstacles and target locations, while the output is the agent's action strategy. The core of reinforcement learning is to obtain feedback through interactions with the environment, i.e., rewards and penalties, to continuously update and optimize the strategy. \cite{kaushik2018learning} propose using the DDPG~\cite{lillicrap2015continuous} algorithm to enable vehicles to learn various driving behaviors in unstructured environments through simulation. Agents improve their behavior through trial-and-error learning, maximizing total rewards by accumulating experience. In path planning applications, DRL effectively handles dynamic, unknown, and complex environments. Agents do not need to pre-know all environmental information; they can find the optimal path from start to target through continuous exploration and learning. 

\cite{liu2020mapper} propose MAPPER, a decentralized partially observable multi-agent path planning method based on evolutionary reinforcement learning, to learn effective local planning strategies in mixed dynamic environments. \cite{semnani2020multi} develop a hybrid algorithm combining deep reinforcement learning and Force-based Motion Planning (FMP) to address distributed motion planning in dense dynamic environments. Their hybrid approach employs the simpler FMP method in stuck, simple, and high-risk situations, and continues to use RL in normal circumstances where FMP cannot generate optimal paths. \cite{wang2023deep} introduce a path planning method for unstructured environments based on deep reinforcement learning, where agents are trained in a low-dimensional simulator constructed from occupancy grid maps. A self-supervised pre-text task helps agents learn environmental knowledge, and an adaptive curriculum learning method improves training efficiency and generalization ability.

Errors in traversability estimation, such as misclassifying soft soil or mud as drivable, can propagate to planning modules and increase the risk of collisions or vehicle immobilization. Future work should explore uncertainty-aware planning and other mitigation strategies to account for perception inaccuracies, enhancing the overall robustness of off-road autonomous navigation.

\section{Motion Control}
\label{control}


In traditional modular autonomous driving systems, the motion control module functions like the ``hands and feet" of the vehicle. Although significant research has been conducted on urban autonomous vehicles, vehicles present a unique set of modeling challenges in unstructured environments. Vehicles navigate through intricate and unpredictable terrains, resulting in complex interactions between the vehicle and the ground. Additionally, these vehicles possess their own intricate dynamics, which adds another layer of complexity. Such challenges can hinder effective high-speed control and planning processes.

\subsection{Challenges of Motion Control in Unstructured Environment}

\begin{itemize} 
	
	\item \textbf{Complex Interactions between Vehicle and Ground}: 
	Vehicles navigate through intricate and unpredictable terrains, such as rugged landscapes, dense forests, and varying soil types, which can change suddenly due to environmental factors like rain or snow. These conditions lead to complex interactions between the vehicle and the ground, including variations in traction, slipping, and uneven load distribution. Such dynamics can significantly affect the vehicle's stability and control, requiring sophisticated algorithms to accurately predict behavior in real time.
	
	\item \textbf{Vehicles Possess Intricate Dynamics}: 
	Each vehicle has distinct characteristics, including weight distribution, suspension design, and tire properties, all of which influence how it responds to inputs and external forces in unstructured environments. For example, a heavy vehicle may behave differently compared to a lighter one when navigating the same terrain, as its momentum and inertia play crucial roles in its movement. Understanding these individual dynamics is essential for developing effective control strategies, as they determine how the vehicle reacts during acceleration, braking, and cornering. Such challenges can hinder effective high-speed control and planning processes, making it critical to account for both the external terrain and the vehicle's internal dynamics in the modeling phase.
	
\end{itemize}

Based on the degrees of freedom of the vehicle, its motion control process can be decoupled into two relatively independent processes: longitudinal control and lateral control. Longitudinal control focuses on precisely managing the vehicle's speed and acceleration by automatically adjusting power or braking actuators, ensuring that the vehicle moves at the desired speed. Lateral control, on the other hand, makes real-time adjustments to the vehicle's direction by regulating the steering mechanism, ensuring accurate tracking of the planned path.
Currently, the main control algorithms used in ground unmanned vehicle motion control modules include Proportional-Integral-Derivative (PID) control, preview tracking control, sliding mode control, and model predictive control (MPC). The current research focus on motion control for autonomous driving in unstructured environments primarily revolves around MPC methods.

\subsection{Proportional-Integral-Derivative (PID) Control}
The PID control algorithm, a classic control method, adjusts the vehicle’s control variables by coordinating proportional, integral, and derivative controllers. Its advantages lie in its simple structure, ease of implementation, and effective control. However, its performance heavily depends on experience-based parameter tuning, and its precision is susceptible to external environmental disturbances.

\subsection{Preview Tracking Control}
The preview tracking control algorithm calculates the vehicle’s desired steering angle by sampling both tracking error and preview path error. It offers strong dynamic adaptability and path-tracking performance but is sensitive to the choice of preview distance, which requires significant engineering experience to determine optimally.

\subsection{Sliding Mode Control}
The sliding mode control algorithm guides the vehicle’s motion state toward the desired state by designing a sliding mode surface, ensuring stable control. This method is robust against system parameter variations and offers a fast response, though it may cause high-frequency oscillations near the sliding surface, potentially affecting the vehicle’s stability and comfort.

\subsection{Model Predictive Control (MPC)}
Model predictive control typically involves developing kinematic and dynamic models of the vehicle that meet various constraints. It combines the current state with predictions of future states over a finite time horizon to solve an optimization problem, seeking the optimal control sequence. While MPC is highly flexible, allowing adjustments to objectives and constraints for different driving scenarios and tasks, it faces challenges such as high computational complexity and a strong dependence on the accuracy of the model.

In complex and extreme off-road environments, vehicles frequently encounter uneven terrain, mixed soft and hard soils, and unpredictable slopes or obstacles, posing significant challenges to traditional fixed-parameter controllers. Adaptive control has therefore become a key research focus. \cite{onozuka2025adaptive} present an online adaptive Model Predictive Control (MPC) framework for off-road vehicles operating on deformable terrains, incorporating a physics-informed, online-adaptable tire model. Simulation results demonstrate that online adaptation significantly improves speed and path-tracking performance when discrepancies exist between assumed and actual terrain parameters.
\cite{gibson2023multi} use a hybrid model with a specially-initialized LSTM for short-term dynamics prediction of an all-terrain vehicle, addressing error accumulation and scalability for sampling-based controllers. By limiting the LSTM to a fixed time horizon, it avoids long-term stability challenges. This flexible approach requires only odometry data for training. In \cite{han2023model}, the system's constraints are parallelized within the MPC, enabling efficient real-time planning and control. A low-level controller ensures the vehicle adapts to varying terrain without prior knowledge, ensuring safe, aggressive driving over challenging unstructured environments.

\subsection{Reinforcement Learning-based Control}
Another class of reinforcement learning (RL)-based controllers focuses on learning strategies to compensate for slip and wheel-terrain interactions on soft surfaces, enabling vehicles to adapt to unknown or extreme terrains. For example, \cite{gupta2024reinforcement} proposes an Actor-Critic reinforcement learning Compensated Model Predictive Controller (AC2MPC) for high-speed off-road driving on deformable terrains, integrating deep RL with MPC to handle unmodeled tire-terrain dynamics. Simulation results demonstrate that the approach outperforms standalone model-based or learning-based controllers across multiple unknown terrains, achieving more accurate speed tracking, faster convergence, and strong generalization with limited training data. These methods highlight the significant potential of adaptive control to enhance vehicle stability and performance in complex terrains, though challenges remain in real-time computational efficiency, perception uncertainty, and cross-terrain generalization.

\subsection{Model Predictive Path Integral (MPPI)}
In off-road autonomous driving, vehicles frequently encounter complex terrains, dynamic obstacles, and highly nonlinear vehicle dynamics, posing significant challenges for traditional optimization- or linearization-based controllers. Recently, Model Predictive Path Integral (MPPI) control, a sampling-based nonlinear optimal control method, has gained widespread attention in off-road environments. MPPI performs stochastic trajectory sampling and cost-weighted averaging directly in the trajectory space, enabling real-time handling of nonlinear, constrained problems without relying on precise vehicle models. This makes it well-suited for rugged terrains, mixed soft and hard soils, and nonlinear tire-terrain interactions.
	
Studies have demonstrated MPPI’s advantages in off-road autonomous navigation. For example, \cite{williams2017model} proposed an MPPI-based off-road planning method that samples trajectories in high-dimensional state space, achieving robust obstacle avoidance and path tracking in unknown or dynamic environments. \cite{xu2025verti} integrated MPPI with vehicle dynamics constraints and terrain information for high-speed off-road driving, showing strong adaptation to soft ground slip and tire-terrain interactions. Furthermore, MPPI’s parallelizable computation allows near real-time operation on GPUs or embedded edge platforms, making it feasible for practical deployment. Despite these strengths, MPPI still faces challenges in sampling efficiency, cost function design, and coordinated planning for multi-vehicle systems. Overall, MPPI-based controllers provide an effective framework for autonomous navigation in complex off-road environments and offer important directions for adaptive and high-robustness control research.

\section{End-to-End Autonomous Driving}
\label{e2e}

\begin{figure} [h]
	\centering
	\includegraphics[width=0.5\textwidth]{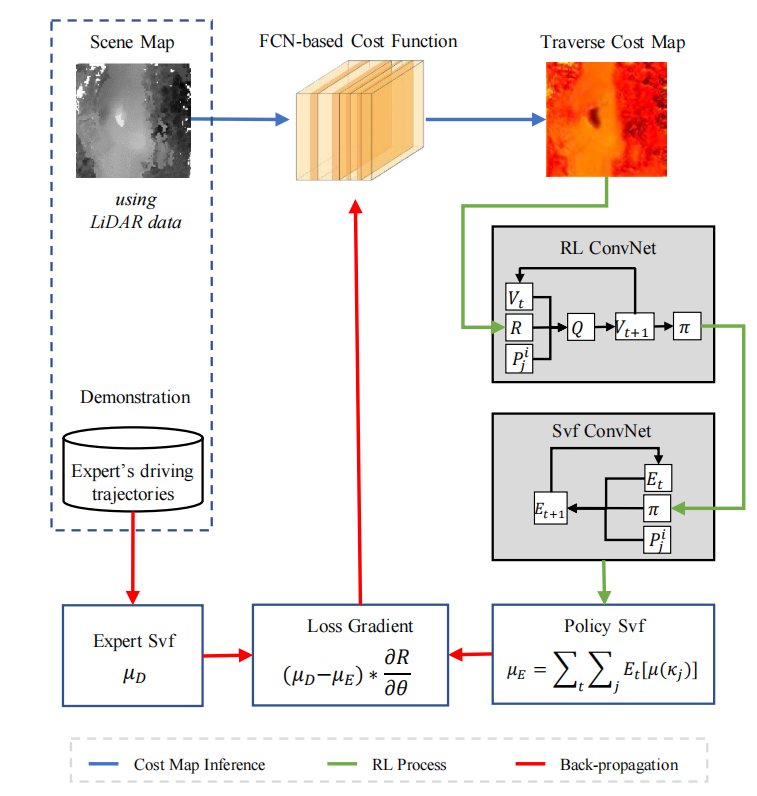}
	\caption{Overview of the end-to-end off-road autonomous vehicles traversability analysis and trajectory
		planning based on deep inverse reinforcement learning~\cite{zhu2020off}.}
	\label{fig:e2e}
\end{figure}
\begin{table}[htbp]
	\caption{End-to-end autonomous driving methods for unstructured environments.}
	\begin{center}
		\resizebox{0.7\textwidth}{!}{
			\begin{tabular}{c | c |c }
				\midrule
				\bf Reference & \bf Algorithm & \bf Type \\ 
				\midrule
				\cite{muller2005off}&Imitation Learning&Behavior Cloning \\
				\cite{leininger2024gaussian} &Imitation Learning&Behavior Cloning \\ 
				\cite{manderson2020learning}  &Imitation Learning &Behavior Cloning    \\ 
				\cite{zhu2020off} &Imitation Learning&Inverse RL\\
				\cite{liang2022adaptiveon} &Online RL &Policy-based\\ 
				\cite{weerakoon2024vapor} &Online RL&Actor-critic\\ 
				\midrule
			\end{tabular}
			\label{tab:e2e}
		}
	\end{center}
\end{table}

The essential issue of autonomous driving lies in exploring and establishing the mapping relationship between sensor observation data and vehicle control commands. Given the inherent complexity of autonomous driving systems, directly characterizing this mapping relationship is extremely challenging. An increasing number of studies have begun to explore the direct application of the data-driven concepts from deep learning to the autonomous driving, thereby giving rise to a novel paradigm for unstructured environments-end-to-end autonomous driving.

End-to-end autonomous driving methods discard the cumbersome task division and rule design found in traditional modular autonomous driving approaches, instead leveraging deep neural networks to directly learn the complex mapping relationship from sensor observation data to vehicle control commands from vast amounts of data~\cite{chen2023end}. The input side of this method can include multimodal information such as visual data, LiDAR point cloud data, platform motion status, and task commands, while the output side mainly generates corresponding control commands for steering mechanisms, power actuators, and brake actuators.

The research origin of end-to-end autonomous driving methods can be traced back to 1988 when Carnegie Mellon University initiated the pioneering ALVINN project~\cite{pomerleau1988alvinn}. In this project, researchers utilized a shallow fully connected neural network to process input images and radar data, achieving end-to-end control of vehicle motion direction. Subsequently, in the DAVE~\cite{muller2005off} project initiated by DARPA, a six-layer neural network was used to map the visual information obtained from stereo cameras into control signals for the steering angle of an unmanned vehicle, successfully realizing end-to-end autonomous obstacle avoidance for the first time. Although these early studies revealed the potential of neural networks in autonomous driving systems to some extent, they still had a significant gap to bridge before achieving complete autonomous driving functionality.

In 2016, NVIDIA further launched the DAVE-2~\cite{bojarski2016end} project based on the DAVE project, successfully achieving end-to-end full-process autonomous driving based on visual information on highways and urban roads through training large-scale deep convolutional neural networks. This groundbreaking achievement not only validated the effectiveness of the end-to-end autonomous driving research paradigm but also truly pushed this research paradigm to the forefront of academia.

With the continuous advancement of deep learning technologies and the ongoing improvement of hardware computing capabilities, end-to-end autonomous driving methods have become a hot research topic for autonomous driving in unstructured environments~\cite{hassan2024pathformer}. Depending on the different learning approaches of the network models, this method has gradually differentiated into two main research directions: imitation learning methods and online reinforcement learning methods, as shown in Table~\ref{tab:e2e}.

\subsection{Challenges of End-to-End Autonomous Driving in Unstructured Environment}


\begin{itemize} 
	
	\item \textbf{Data Scarcity}: 
	In the training process of end-to-end autonomous driving systems, the quality and quantity of data are crucial. Currently, data on unstructured environments is relatively scarce, primarily due to the complexity and diversity of these environments, which makes data collection challenging. For example, various scenarios like urban streets, rural roads, or forest paths may contain different obstacles, pedestrians, and traffic signs, each with unique dynamic characteristics. However, the lack of effective data collection mechanisms and standards often results in existing datasets being insufficient to cover this diversity, leading to biases in model training. Additionally, the variability in weather and lighting conditions in unstructured environments requires datasets to include scenes under various conditions to ensure the model's robustness and adaptability. Thus, data scarcity becomes a significant bottleneck in the development of end-to-end autonomous driving technology.
	
	\item \textbf{Lack of Simulation Environments}: 
	Effective training and testing require a high-quality simulation environment, especially for unstructured environments. However, there is currently a relative lack of simulation platforms specifically designed for these settings. Most existing simulation tools focus mainly on structured urban roads or simple traffic scenarios, limiting their applicability in complex and dynamic conditions. The diversity of unstructured environments, including irregular terrains, sudden obstacles, and unpredictable weather changes, makes it particularly challenging to design a comprehensive simulation system. Furthermore, current simulation systems often lack sufficient realism, making it difficult to accurately simulate human behavior and traffic conditions, which leads to subpar performance of trained models in the real world. This lack of simulation environments further restricts the advancement and application of end-to-end autonomous driving technology in unstructured settings.
	
\end{itemize}

\subsection{Imitation Learning}
Imitation learning is a supervised learning paradigm that trains agents using recorded expert behavior patterns, aiming for the agents to accurately emulate the relevant behavioral strategies of experts and achieve optimal performance. In end-to-end autonomous driving methods based on imitation learning, human drivers are positioned as experts, and specific actions they perform during driving, such as steering, acceleration, and braking, are regarded as expert behavior patterns for ground unmanned platforms to learn and imitate. During human driving, data collected by onboard sensors is recorded as state information, while the corresponding control commands are recorded as action information. By utilizing all recorded state-action pairs, an expert driving dataset can be constructed. 
Based on specific learning objectives, these methods can be further subdivided into two subcategories: behavior cloning-based methods and inverse reinforcement learning-based methods. 

\subsubsection{Behavior Cloning}

Behavior cloning-based end-to-end autonomous driving methods train network models by minimizing the differences between human driving behavior strategies and those of ground unmanned platforms. This enables the platforms to simulate driving behavior strategies that closely resemble those of human drivers. 
\cite{leininger2024gaussian} propose a geometric-based framework for navigating uneven terrain without relying on pre-existing maps. This framework integrates a Sparse Gaussian Process (SGP) local map with a RRT* planner. The process begins with creating a high-resolution SGP local map, which provides an interpolated depiction of the robot's immediate environment. This map captures key environmental features such as height variations, uncertainties, and slope characteristics. Next, a traversability map is generated from the SGP representation to guide the planning process. The RRT* planner then efficiently creates real-time navigation paths, avoiding untraversable terrain and directing the robot towards its goal. By combining SGP-based terrain interpretation with RRT* planning, ground robots can safely navigate areas with varying elevations and steep obstacles. The proposed approach's effectiveness is validated through rigorous simulation testing, demonstrating notable improvements in safe and efficient navigation over existing methods.

Behavior cloning-based methods have demonstrated advantages of simplicity and efficiency during implementation, while avoiding the cumbersome design requirements of reward mechanisms. However, such methods also generally face a series of challenging issues, such as dependence on large-scale training data, insufficient generalization ability, distribution shifts, and causal confusion, making them difficult to effectively cope with complex, unknown environments and various edge cases.

\subsubsection{Inverse Reinforcement Learning}
End-to-end autonomous driving methods based on inverse reinforcement learning aim to learn implicit reward functions that can explain human driving behavior using expert driving datasets. These reward functions are viewed as the underlying motivations driving human experts to perform specific driving behaviors. With the acquired reward functions, ground unmanned platforms can infer and execute driving behaviors consistent with expert actions. 

As shown in Fig.~\ref{fig:e2e}, \cite{zhu2020off} introduce a method for off-road traversability analysis and trajectory planning using Deep Maximum Entropy Inverse Reinforcement Learning. Addressing the challenge of managing the exponential increase in state-space complexity while incorporating vehicle kinematics, the approach employs two convolutional neural networks: RL ConvNet and Svf ConvNet. These networks encode kinematic information into convolutional kernels, enabling efficient forward reinforcement learning. Experimental validation is conducted in off-road environments where scene maps are generated using 3D LiDAR data. Expert demonstrations consist of real vehicle driving trajectories in the scene or synthesized behaviors representing specific actions, such as traversing negative obstacles.

The work in~\cite{manderson2020learning} employs a hybrid model-based and model-free reinforcement learning technique that is fully self-supervised for labeling terrain roughness and predicting collisions using onboard sensors. Notably, it integrates both first-person and overhead aerial images as inputs to the model. The fusion of these complementary inputs enhances planning foresight and fortifies the model against visual obstructions. Experimental results demonstrate its capability to generalize effectively across environments rich in vegetation, diverse rock formations, and sandy trails.

Compared to behavior cloning-based methods, inverse reinforcement learning-based methods exhibit stronger adaptability to complex environments and offer a certain degree of interpretability. However, these methods also face challenges such as reliance on large-scale training data, high computational complexity, instability during the training process, and limited generalization capabilities.

\subsection{Online Reinforcement Learning}
End-to-end autonomous driving methods based on online reinforcement learning maximize the accumulated rewards over time through continuous interaction and exploration between ground unmanned platforms and their environments, thereby learning optimal driving behavior strategies. These methods do not rely on manually designed rules or the pre-collection of large amounts of expert driving data, and they are gradually being applied in end-to-end autonomous driving tasks. Depending on the learning approach, these methods can be further subdivided into several subcategories, including value-based methods, policy-based methods, and actor-critic methods.

\subsubsection{Value-based Methods}

Value-based methods learn a value function through the interactive exploration process between the ground unmanned platform and the environment. The value function accurately assesses the expected return from executing a specific driving behavior strategy in a given state, allowing the platform to select the optimal driving behavior strategy by comparing the differences in expected returns. For example, \cite{wolf2017learning} propose an end-to-end autonomous driving framework based on Deep Q-Networks (DQN), which uses monocular images as input and learns human driving behavior strategies through a designed action-based reward function, ultimately achieving autonomous driving performance comparable to that of human drivers in a simulated environment. \cite{min2018deep} design a reward function that comprehensively considers factors such as speed, safety, and lane-changing frequency, and also utilized DQN to achieve autonomous driving in a highway simulation environment. 

\subsubsection{Policy-based Methods}

Policy-based methods directly learn the probability distributions of various driving behavior strategies to be executed in a given state during the interactive exploration process, favoring the execution of the strategy with the highest probability. 
\cite{liang2022adaptiveon} introduce an outdoor navigation algorithm that ensures stable, efficient robot guidance. Using a multi-stage training pipeline with Proximal Policy Optimization (PPO), the method addresses drift, stability, and collision avoidance on complex terrains. It incorporates generalized environmental parameters and LiDAR data from a Unity simulator to bridge the sim-to-real gap. Evaluated on Clearpath Husky and Jackal robots, this approach effectively achieves robust navigation in both simulated and real-world settings.

\subsubsection{Actor-critic Methods}
Actor-critic methods combine the core ideas of the above two approaches by simultaneously learning the value function and policy function through continuous interaction and exploration between the ground unmanned platform and the environment. In this process, the value function acts as a critic, assessing the expected return from executing specific driving behavior strategies in the current state, while the policy function acts as an actor, selecting appropriate driving behavior strategies based on the current state. VAPOR~\cite{weerakoon2024vapor}, a novel method for autonomous navigation of legged robots in dense outdoor vegetation using offline Reinforcement Learning. This approach employs an actor-critic network to train a policy on data from real environments, utilizing cost maps from 3D LiDAR and proprioceptive data. The policy learns about obstacle properties like height and stiffness. A context-aware planner then adjusts the robot's velocity based on potential hazards such as entrapment and narrow passages.

\section{Multi-Agent Collaborative Systems}
\label{multi}

In recent years, Multi-Agent Collaborative Systems (MACS) have emerged as a research hotspot in the field of off-road autonomous driving~\cite{wang2025multi}. Compared with single-vehicle autonomous navigation, multi-vehicle (or multi-robot) collaborative systems achieve more efficient environmental perception, path planning, and decision-making through information sharing and task allocation, making them particularly suitable for unstructured scenarios such as disaster search and rescue, mining transportation, and agricultural operations. In this context, \cite{dinneweth2022multi} conduct a comprehensive survey on multi-agent reinforcement learning methods for cooperative driving, analyzing key challenges such as communication constraints and distributed policy optimization in collaborative perception and decision-making. \cite{zhang2022h2gnn} propose a distributed multi-robot cooperative exploration framework based on Graph Neural Networks (GNNs), enabling robots to achieve globally efficient mapping in complex and rugged terrains. The CTU-CRAS-NORLAB team developed a multi-robot exploration framework for ground robots, focusing on planning, traversability estimation, and decentralized coordination in large-scale, communication-limited underground environments during the 2021 DARPA Subterranean Challenge~\cite{bayer2023autonomous}. The framework combines dense local mapping with a shareable sparse topometrical map, enabling fully autonomous exploration and effective multi-robot coordination, as demonstrated in both Virtual and Systems tracks of the competition.

\section{Datasets}
\label{data}

\begin{figure*} [h]
	\centering
	\includegraphics[width=0.9\textwidth]{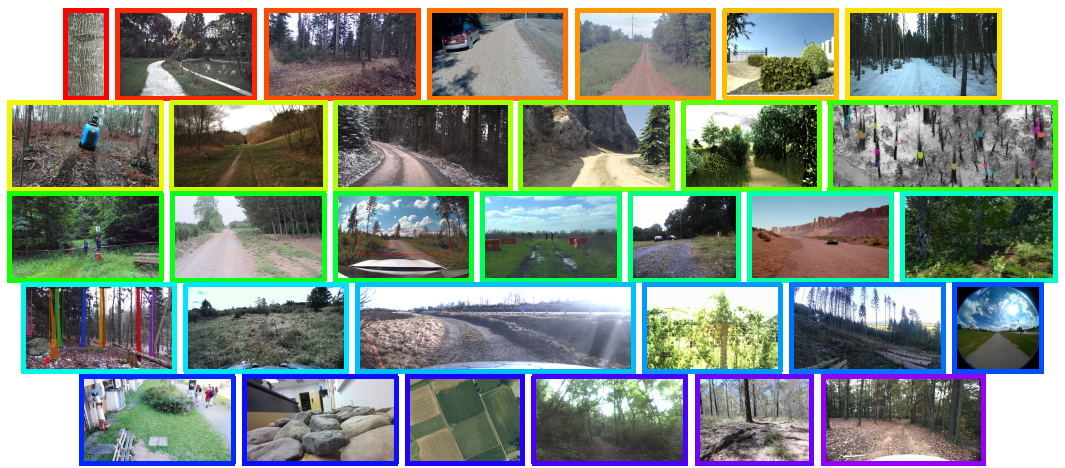}
	\caption{An overview of the public autonomous driving datasets of unstructured environments~\cite{mortimer2024survey}.}
	\label{fig:data_pic}
\end{figure*}

\begin{figure*} [h]
	\centering
	\includegraphics[width=0.98\textwidth]{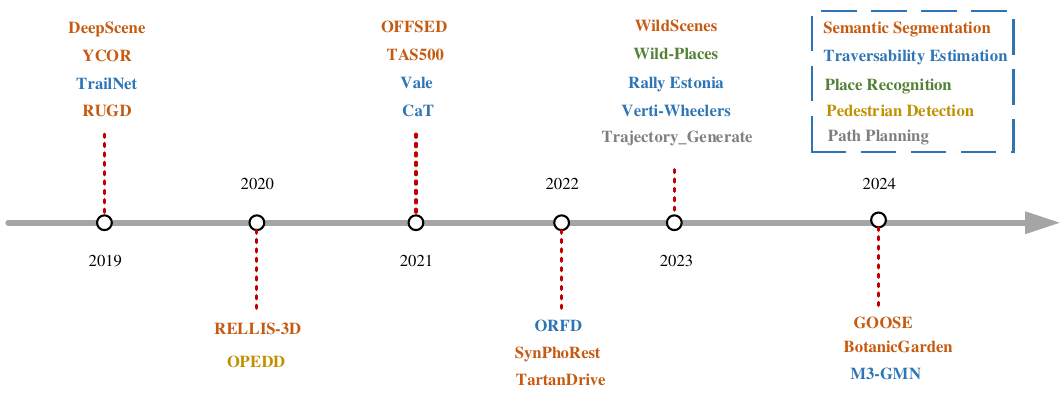}
	\caption{Autonomous driving datasets for unstructured environments.}
	\label{fig:datasets}
\end{figure*}

In recent years, data-driven deep learning algorithms have significantly advanced autonomous driving technology. Popular autonomous driving datasets such as KITTI~\cite{kitti}, Waymo~\cite{waymo}, and nuScenes~\cite{nuscenes} primarily focus on structured scenes. Recently, there has been increasing attention on publicly releasing autonomous driving datasets for unstructured environments, as shown in Fig.~\ref{fig:data_pic}. As illustrated in Fig.~\ref{fig:datasets} and Table~\ref{ts_dataset}, according to target tasks, these datasets can be categorized into traversability estimation, semantic segmentation, and other purposes.

\begin{table*}
	\caption{Details of datasets for unstructured environments.}
	\begin{center}
		\setlength{\tabcolsep}{0.8mm}{
			\begin{tabular}{c|c|c|c|c}
				\midrule
				\bf Task&\bf Dataset &\bf Sensor & \bf \# Frames
				& \bf \# Classes \\ 
				\midrule
				\multirow{5}*{{Traversability Estimation}}&TrailNet&RGB Camera &- &-  \\
				&Vale&RGB Camera &600 &4 \\
				&CaT&RGB Camera & 3,624 &- \\
				&ORFD& RGB Camera, LiDAR &12,128 &3  \\
				&Verti-Wheelers&Stereo Camera &70,143 &- \\
				&M3-GMN&LiDAR, RGB Camera, GPS/INS &1578 &- \\
				&ORAD-3D&LiDAR, RGB Camera&57,808 &- \\
				\midrule
				\multirow{10}*{{Semantic Segmentation}}&DeepScene&RGB, NIR, Stereo&366 &6  \\
				&YCOR&RGB Camera & 1,076 &8 \\
				&RUGD&RGB Camera&7,546 & 24  \\	
				&TAS500&RGB Camera&640 &24  \\
				&RELLIS-3D&RGB, Stereo Camera &6,235 &20 \\
				&BotanicGarden&RGB, Stereo Camera &1,181 &27  \\
				&GOOSE& RGB, NIR Camera&8,790 & 64  \\
				&WildScenes&RGB Camera, LiDAR & 12,148 &15  \\
				&ORAD-3D&LiDAR, RGB Camera&57,808 &8 \\
				\midrule
				Pedestrian Detection &OPEDD&Stereo Camera&1,018 &- \\
				\midrule
				Dynamics Models &TartanDrive&Stereo Camera&200,000 &- \\
				\midrule
				End-to-end Driving &Rally Estonia& RGB Camera, LiDAR &- &-  \\
				&ORAD-3D&LiDAR, RGB Camera&57,808 &- \\
				\midrule
				Place Recognition &Wild-Places&LiDAR & 63,000 &-  \\
				\midrule
				Path Planning&Trajectory\_Generate&LiDAR, RGB-D Camera, GNSS& 20,859 &-  \\
				
				\midrule
			\end{tabular}
			\label{ts_dataset}
			}
	\end{center}
\end{table*}

\subsection{Traversability Estimation}
For autonomous vehicles, the foremost aspect of environmental perception is identifying the traversable area.
TrailNet~\cite{hoveidar2018autonomous} is among the pioneering datasets that examine various road surface types and make the recorded camera data publicly available.
The Vale~\cite{hosseinpoor2021traversability} dataset, recorded from a drone, was annotated for traversability levels based on surface height differences. These levels correlate with robot locomotion methods like wheeled, tracked, and legged.
The CaT~\cite{sharma2022cat}, akin to Vale, supports off-road autonomous driving with high-resolution images, LiDAR data, and detailed terrain annotations, enhancing traversability analysis and autonomous driving performance in off-road conditions.

Recent datasets have expanded to over 10,000 frames, promoting research on traversability estimation. The ORFD~\cite{orfd} dataset provides high-resolution images and annotations for various off-road scenarios, helping to identify navigable areas. The benchmark evaluates different approaches to freespace detection, promoting the development of more accurate and robust algorithms.
Verti-Wheelers~\cite{datar2023toward} addresses the challenges of wheeled mobility on vertically challenging terrain. The authors present new platforms, datasets, and algorithms designed to enhance the capabilities of wheeled robots in environments with significant vertical obstacles. The datasets include various types of challenging terrains, while the proposed algorithms focus on navigation, stability, and control.
M3-GMN~\cite{m3gmn} introduces a diverse dataset designed to enhance grid map-based navigation for autonomous vehicles. This dataset spans multiple environments, LiDAR types, and tasks, capturing both structured and unstructured environments. It includes a range of challenging scenarios, such as moving objects, negative obstacles, steep slopes, cliffs, and overhangs.

ORAD-3D~\cite{orad} establishes a comprehensive suite of benchmark evaluations spanning five fundamental tasks: 2D free-space detection, 3D occupancy prediction, rough GPS-guided path planning, vision-language model-driven autonomous driving, and world model for off-road environments. Together, the dataset and benchmarks provide a unified and robust resource for advancing perception and planning in challenging off-road scenarios.

\subsection{Semantic Segmentation}
In recent years, the development of deep learning technologies has enabled the application of semantic segmentation algorithms to unstructured environments. Several autonomous driving datasets focused on semantic segmentation in unstructured scenes have been made publicly available, providing researchers with valuable resources for their studies.

We will start by introducing datasets for semantic segmentation that contain only a few hundred frames of images.
DeepScene~\cite{valada2017deep} explores multispectral imagery and multimodal data fusion to enhance semantic scene understanding in forested environments, aiding forestry monitoring, environmental management, and autonomous vehicle navigation.
OFFSED~\cite{offsed} targets off-road semantic segmentation tasks, providing high-resolution images with fine-grained semantic labels across various off-road scenarios, including forests, fields, and unpaved paths.
TAS500~\cite{metzger2021fine} offers a fine-grained dataset for unstructured driving scenarios, with detailed annotations of terrain types and obstacles in off-road trails and rural paths, enhancing semantic segmentation algorithm performance.

There are also some datasets with over a thousand frames.
YCOR~\cite{maturana2018real} features 1,076 images from Western Pennsylvania and Ohio, collected across three seasons, with polygon-based labeling into eight classes: sky, rough trail, smooth trail, traversable grass, high vegetation, non-traversable low vegetation, and obstacles.
SynPhoRest~\cite{nunes2022procedural} includes RGB images with pixel-perfect segmentation maps, depth maps, and point clouds with point-wise labeling, supporting tasks like semantic segmentation, object detection, and forest navigation.
BotanicGarden~\cite{liu2024botanicgarden} is designed for robot navigation in natural environments, providing high-quality images and detailed annotations of botanical gardens to improve navigation algorithms in complex settings.

Recently, some datasets with over 5,000 frames have been introduced, advancing research in semantic segmentation for unstructured environments.
RUGD~\cite{rugd} contains a wealth of images and sensor data used to train and evaluate perception and navigation algorithms for autonomous vehicles.
RELLIS-3D~\cite{rellis3d} is a multi-modal dataset specifically designed for off-road robotics. It includes various sensor data, such as LiDAR, cameras, and IMUs, which are essential for tasks like 3D scene understanding, navigation, and obstacle avoidance. 
The GOOSE~\cite{mortimer2023goose} dataset includes diverse terrains and obstacles, aiming to support the development of robust perception algorithms for tasks such as semantic segmentation, object detection, and navigation in challenging environments.
WildScenes~\cite{wildscenes} is a bimodal benchmark dataset that encompasses multiple extensive traversals within natural environments. It provides semantic annotations in high-resolution 2D images, dense 3D LiDAR point clouds, and accurate 6-DoF pose information.

\subsection{Other Datasets}
There are also datasets designed for autonomous driving in unstructured environments, focusing on tasks such as pedestrian detection, dynamics models, end-to-end driving, and location recognition.

OPEDD~\cite{neigel2020opedd} is a dataset focused on pedestrian detection in off-road environments. Compared to urban settings, pedestrian detection in off-road conditions faces more challenges, such as complex backgrounds and irregular lighting conditions.

TartanDrive~\cite{triest2022tartandrive} is a large-scale dataset aimed at learning off-road dynamics models. It includes extensive sensory data from off-road driving scenarios, capturing the dynamic interactions between the vehicle and various types of terrain. 

Rally Estonia~\cite{tampuu2023lidar} explores the use of LiDAR data as a substitute for camera images in end-to-end driving systems. 

Wild-Places~\cite{wideplaces} is aimed at LiDAR place recognition in unstructured natural settings. It features eight lidar sequences gathered using a handheld sensor payload over fourteen months, containing 63,000 undistorted lidar submaps and accurate 6DoF ground truth data.

Trajectory\_Generate~\cite{uzawa2023dataset} uses an RGB-D camera combined with an IMU to collect data for image-based end-to-end path planning. The data was collected from three different locations: a university greenhouse, a university park, and a nearby natural trail.
\section{Future Perspectives}
\label{future}

To advance the development of autonomous driving technology in unstructured environments, future research should focus on the following core areas:

\textbf{High-Precision Dynamic Map Construction.}
Develop high-resolution offline maps that capture terrain details and dynamic environmental changes using LiDAR and photogrammetry~\cite{zhu2021deep,wei2021aa,wei2022bidirectional}. Emphasize spatiotemporal modeling to address seasonal and weather variations by integrating time-series data (e.g., monthly point clouds) and employing techniques such as 4D Gaussian processes or spatiotemporal CNNs. A digital twin system with dynamic updates can provide a high-precision, temporally-aware knowledge base to support navigation, path planning, and decision-making in complex, changing terrains.

\textbf{Multimodal Pose Estimation System.}
To address the challenge of ambiguous terrain features, new algorithms for camera-LiDAR-IMU multi-sensor fusion need to be developed. Using deep learning methods to analyze complex spatial relationships, a real-time positioning framework based on probabilistic reasoning should be created to solve pose estimation issues in feature-deficient scenarios, thereby improving system robustness in dynamic environments.

\textbf{Environmental Perception Enhancement.}
Build a multimodal perception network combining vision, LiDAR, and RADAR ~\cite{li2024pre}, leveraging graph neural networks and Transformer architectures to achieve fine-grained semantic understanding of natural objects. The focus should be on enhancing open-set recognition in unstructured environments and developing explainable perception models~\cite{shaban2022semantic} to enable adaptive classification and risk prediction for unknown obstacles.

\textbf{Intelligent Path Planning and Decision Making.}
Develop a dynamic path planning framework based on reinforcement learning, integrating real-time terrain scanning and predictive environmental modeling. By designing cost functions that account for risk perception, a multi-objective decision system~\cite{wang2024motion} should be created to optimize both safety and energy consumption, enabling collaborative optimization of global path planning and local obstacle avoidance in complex terrains.

\textbf{Nonlinear Motion Control Architecture.}
Develop an adaptive controller with terrain prediction capabilities, and create a hybrid control strategy that combines model predictive control and learning methods. The focus should be on addressing dynamic coupling issues in extreme conditions, such as steep slopes and soft ground, while establishing a slip compensation mechanism that considers tire-ground interaction characteristics, improving motion stability in rapidly changing terrain.

\textbf{End-to-End System Integration and Innovation.}
Create a neural radiance field simulation environment for off-road scenarios and develop an end-to-end driving architecture based on world models. By combining imitation learning and reinforcement learning, the system should be capable of making end-to-end decisions from raw sensor inputs to control outputs, minimizing reliance on manually defined rules.

\textbf{Multidimensional Dataset Construction.}
Existing unstructured off-road datasets remain limited in scale and diversity, highlighting the need to expand their richness in future work. Current datasets also provide little analysis of semantic ambiguities, such as distinguishing between passable grass and dense tall vegetation, which should be addressed to improve perception models. Moreover, most existing datasets focus on static environments, while dynamic elements—such as swaying branches or flowing water—are rarely captured, limiting their applicability for training dynamic obstacle avoidance algorithms. Future efforts should therefore aim to collect datasets that include both dynamic elements and terrain–vehicle interactions.

\textbf{Extreme Environmental Conditions.}
Future research should systematically evaluate the performance of off-road autonomous driving algorithms under extreme environmental conditions, such as heavy rain, snow, or dynamic obstacles like wildlife and falling rocks. For instance, the impact of LiDAR point cloud noise caused by rain on traversability estimation accuracy remains largely unexplored. Investigating sensor robustness, perception uncertainty, and adaptive planning strategies under such conditions will be critical for achieving reliable and safe off-road autonomous navigation in real-world scenarios.

\textbf{Hardware Heterogeneity.}
Future studies should investigate the impact of heterogeneous hardware configurations on the performance of off-road autonomous driving algorithms. For example, differences between low-resolution (16-beam) and high-resolution (128-beam) LiDAR, or between consumer-grade and high-end IMUs, can significantly affect perception accuracy, localization, and control. Quantifying these effects will help researchers and practitioners select appropriate hardware for specific scenarios and optimize system performance under resource constraints.

\textbf{Edge Computing Integration.}
Off-road autonomous driving in unstructured environments faces challenges such as weak networks, high latency, and large sensor data~\cite{liu2019edge}. Traditional cloud computing often fails to meet real-time demands. Future work should leverage edge computing for local and collaborative sensor processing, including LiDAR compression, semantic extraction, and dynamic object recognition. Key directions include task allocation, computation offloading, robustness under limited connectivity, and scalable multi-vehicle coordination. Progress in these areas is essential for fully autonomous, resilient off-road operation.

\textbf{Explainable Research for Off-Road AD.}
Future work should develop explainable perception and decision-making methods for unstructured off-road environments, where complex terrain, dynamic obstacles, and sensor noise challenge autonomy. Techniques that highlight which LiDAR points, camera features, or fused sensor data drive obstacle detection, traversability estimation, and planning decisions are essential to enable real-time validation, build operator trust, and ensure safety in critical applications such as military missions and disaster response.
\section{Conclusion}
\label{conclusion}
Due to the diversity and complexity of natural environments such as rural roads, mountainous terrains, and wilderness areas, autonomous driving in unstructured environments poses numerous challenges. These environments lack standardized road markings and are susceptible to unpredictable obstacles like wildlife and dynamic weather conditions. This review provides an overview of current advancements in autonomous driving in unstructured environments, covering areas such as offline mapping, pose estimation, environmental perception, path planning, motion control, end-to-end driving, and datasets. It serves to facilitate a rapid understanding of the research progress in this field. While achieving fully autonomous driving in unstructured environments presents significant hurdles, ongoing research and technological advancements continue to pave the way for safer, more adaptive autonomous vehicles capable of navigating the world’s most challenging terrains.

\subsubsection*{Acknowledgments}
This work was supported by National Natural Science Foundation of China under Grant No.U23B2034, No.62203424 and No.62176250, and the Innovation Program of Institute of Computing Technology, Chinese Academy of Sciences under Grant No. 2024000112.

\bibliographystyle{apalike}
\bibliography{jfrExampleRefs}

\end{document}